\documentclass[sigconf]{acmart}
\usepackage{amsfonts}
\usepackage{algorithm}  
\usepackage{algpseudocode}  
\usepackage{amsmath}  
\usepackage{diagbox}
\usepackage{multirow}
\usepackage{diagbox}
\usepackage{subfigure}
\usepackage{graphicx}
\usepackage{stfloats}

\AtBeginDocument{%
  \providecommand\BibTeX{{%
    \normalfont B\kern-0.5em{\scshape i\kern-0.25em b}\kern-0.8em\TeX}}}

\setcopyright{acmcopyright}
\copyrightyear{2022} 
\acmYear{2022} 
\setcopyright{acmcopyright}\acmConference[WWW '22]{Proceedings of the ACM Web Conference 2022}{April 25--29, 2022}{Virtual Event, Lyon, France}
\acmBooktitle{Proceedings of the ACM Web Conference 2022 (WWW '22), April 25--29, 2022, Virtual Event, Lyon, France}
\acmPrice{15.00}
\acmDOI{10.1145/3485447.3512211}
\acmISBN{978-1-4503-9096-5/22/04}

\begin{document}

\title{Dual Space Graph Contrastive Learning}

\author{Haoran Yang}
\authornote{Both authors contributed equally to this research.}
\email{haoran.yang-2@student.uts.edu.au}
\affiliation{%
	\institution{University of Technology Sydney}
	\city{Sydney}
	\state{NSW}
	\country{Australia}
}

\author{Hongxu Chen}
\authornotemark[1]
\authornote{Corresponding author.}
\email{hongxu.chen@uts.edu.au}
\affiliation{%
	\institution{University of Technology Sydney}
	\city{Sydney}
	\state{NSW}
	\country{Australia}
}

\author{Shirui Pan}
\email{shirui.pan@monash.edu}
\affiliation{%
	\institution{Monash University}
	\city{Melbourne}
	\state{VIC}
	\country{Australia}
}

\author{Lin Li}
\email{cathylilin@whut.edu.cn}
\affiliation{%
	\institution{Wuhan University of Technology}
	\city{Wuhan}
	\state{Hubei}
	\country{China}
}

\author{Philip S. Yu}
\email{psyu@cs.uic.edu}
\affiliation{%
	\institution{University of Illinois at Chicago}
	\city{Chicago}
	\state{Illinois}
	\country{U.S.A}
}

\author{Guandong Xu}
\authornotemark[2]
\email{guandong.xu@uts.edu.au}
\affiliation{%
	\institution{University of Technology Sydney}
	\city{Sydney}
	\state{NSW}
	\country{Australia}
}

\renewcommand{\shortauthors}{Yang and Chen, et al.}

\begin{abstract}

Unsupervised graph representation learning has emerged as a powerful tool to address real-world problems and achieves huge success in the graph learning domain. Graph contrastive learning is one of the unsupervised graph representation learning methods, which recently attracts attention from researchers and has achieved state-of-the-art performances on various tasks. The key to the success of graph contrastive learning is to construct proper contrasting pairs to acquire the underlying structural semantics of the graph. However, this key part is not fully explored currently, most of the ways generating contrasting pairs focus on augmenting or perturbating graph structures to obtain different views of the input graph. But such strategies could degrade the performances via adding noise into the graph, which may narrow down the field of the applications of graph contrastive learning. In this paper, we propose a novel graph contrastive learning method, namely \textbf{D}ual \textbf{S}pace \textbf{G}raph \textbf{C}ontrastive (DSGC) Learning, to conduct graph contrastive learning among views generated in different spaces including the hyperbolic space and the Euclidean space. Since both spaces have their own advantages to represent graph data in the embedding spaces, we hope to utilize graph contrastive learning to bridge the spaces and leverage advantages from both sides. The comparison experiment results show that DSGC achieves competitive or better performances among all the datasets. In addition, we conduct extensive experiments to analyze the impact of different graph encoders on DSGC, giving insights about how to better leverage the advantages of contrastive learning between different spaces.
\end{abstract}



\maketitle

\section{Introduction}

Graph neural networks (GNNs) \cite{gcn, gat, graphsage, gin, pme} leverage the expressive power of graph data in modeling complex interactions among various entities under real-world scenarios including social networks \cite{gcn} and molecules \cite{molecules}. GNNs are able to process variable-size graph data and learn low-dimensional embedding via an iterative process of transferring and aggregating the semantics from topological neighbors. They have achieved huge success on various tasks. GNNs require a large amount of manually labelled data to learn informative representations via the supervised learning protocol. However, it is not always realistic to obtain enough effective labels due to expensive labour work and privacy concerns in some context. Thus, unsupervised approaches such as reconstruction based methods \cite{matrix-completion} and graph contrastive learning methods \cite{gcc, graph-aug, mvgcl} are coupled with GNNs to tackle the problem in a self-supervised manner. Recently, graph contrastive learning has emerged as a successful method for unsupervised graph representation learning and achieved state-of-the-art on various tasks including node classification, graph classification as well as applications in recommender systems \cite{hmg-cr, hgcr} and drug discovery \cite{drug-discovery}.

The key idea of graph contrastive learning is to construct proper contrasting pairs and try to learn the underlying important structural semantics of the input graph. However, existing approaches to generate contrasting pairs are quite limited, the majority of these methods are graph perturbations including node dropping, edge perturbation and attribute masking \cite{graph-aug, gca}. As a result, these methods introduce noisy signals to produce corrupted views, which degrade the model performance \cite{mvgcl}. Therefore, it is always a confronting challenging problem to design effective ways to generate contrasting pairs. Let us look closer, what graph contrastive learning actually does is that it pushes negative pairs far away from each other and make positive pairs as similar as possible to each other. For negative pairs, it is easier to distinguish them against irrelevant graphs derived from other datasets \cite{gcc} or other irrelevant graphs in the same training batch \cite{graph-aug}, which will not involve significant noises. In contrast, the positive samples are usually derived from the input graph via graph perturbation methods, and the noise is inevitable. To avoid introducing too much noise, many researchers instead adopt sub-graph sampling to generate positive samples \cite{gcc, graph-aug, s2grl}. The intuition is that each sub-graph has unique focus on certain aspects of the graph semantics and different sub-graphs or local structures can hint the full spectrum of the semantics \cite{graph-aug} carried on the graphs. Nevertheless, how to generate sufficient views containing unique and informative features is still an open problem. 


Inspired by recent progress of geometric graph mining in hyperbolic space \cite{hgcn, hyperbolic-geometry}, which has achieved satisfactory results under various real-world scenarios, e.g, recommendation systems \cite{hyperbolic-seq-rec}, we innovatively utilize the greater expresiveness of different embedding spaces including the hyperbolic space and the Euclidean space to generate multiple views of the input graph. There are two major advantages of the hyperbolic space. The first is that the hyperbolic space takes smaller space to accommodate a given graph with complex structures, which means hyperbolic embedding has unique expressiveness compared to its counterpart in Euclidean space. 
The most obvious uniqueness is that Euclidean space expands polynomially, while the hyperbolic space expands exponentially. 
Therefore, we can leverage that with much lower dimensional embeddings to represent a graph, which will result in purer, compact but powerful embedding spaces. According to the experimental results that showcase the impact of hidden dimensions on performances of our proposed DSGC framework, it is shown that the our model achieves stable and better performances within smaller dimensional hyperbolic embedding spaces. The second advantage is that the hyperbolic space is more capable of capturing hierarchical structures \cite{hyperbolic-geometry, embedding-space} exhibited in graph data, which implies that hyperbolic space is also suitable to be used as a special view for graph contrastive learning paradigm.

Despite the advantages of hyperbolic space embedding, one cannot ignore the power of the traditional Euclidean space \cite{gcn, gat, graphsage, gin}. Compared to the hyperbolic space, vector calculation in Euclidean space is more efficient as distance metrics in Euclidean space does not need inverse trigonometric functions. Moreover, metrics in Euclidean space is able to differentiate the relative distances among the data points (i.e., satisfying the triangle inequality), and the superiority is evidenced by various famous clustering methods, including k-NN \cite{knn} and HDBSCAN \cite{hdbscan}. Thus, introducing both the hyperbolic and Euclidean spaces together to generate distinct views could illustrate the hierarchical topology and relative distances among the nodes in the original graph, which are beneficial to graph contrastive learning. To fully leverage such advantages, we propose to select different graph samplers to sample sub-graphs in different spaces. Here we argue that random sub-graph extraction cannot fully leverage the advantages of contrasting between Euclidean space and the hyperbolic space. As mentioned previously, different spaces have different advantages, it is better to fully leverage both of them. 
Hence, we should have different strategies to sample sub-graphs. For Euclidean space, we need to sample a sub-graph that is able to demonstrate the skeleton of the input graph (i.e., illustrating relative locations among nodes). For hyperbolic space, a sub-graph carrying hierarchical information and topology structure is expected. 
Further, we note that all the current works \cite{gcc, graph-aug, mvgcl, s2grl} on graph contrastive learning adopt the same graph encoder to encode different views. We question that such a strategy will limit the distinctions among various views, which may undermine the performances of graph contrastive learning. To avoid this, we also propose different graph encoders for different views. 

The main contributions of this paper are summarized as follows:
\begin{itemize}
    \item We first utilize graph representations in different spaces to construct contrsting pairs, which could leverages the advantages from both the hyperbolic space and the Euclidean space.
    \item We innovatively propose to select different graph encoders for representation learning in different spaces, which is also verified by the experiemnts as feasible way to generate different views of the input graph.
    \item We conduct extensive experiments to show the superiority of the proposed DSGC on all the datasets comparing to all the baselines. Detailed study is executed to analyze the strategy about selecting different graph encoders for the graph representation learning in different spaces. We also illustrate if the model can achieve better performances with lower hidden dimensions to verify whether DSGC leverages the advantages of the hyperbolic space.
\end{itemize}

\section{Related Work}

\subsection{Graph Contrastive Learning}
Graph contrastive learning methods are currently state-of-the-art in unsupervised graph representation learning. Deep Graph Infomax (DGI) \cite{dgi} first introduced Deep Infomax \cite{dim} into graph learning and achieve satisfying results via maximize the mutual information between local structure and global context. Some other works generalized such a strategy and try to measure the mutual information between different instances. For example, GCC \cite{gcc} conducted contrastive learning among different sub-graphs extracted from different input graphs, which can be regarded as the mutual information estimation among different local structures. GraphCL \cite{graph-aug} is another type, which are different from GCC. It maximized mutual inforamtion between the input graph and the augmented graph, which is contrasting between two different global views. We note that most current graph contrastive learning methods including \cite{mvgcl, gca, gmi} all conduct contrastive learning between corrupted or augmented graphs or sub-graphs. The methods to generate different view of the input graph for contrastive learning are not rich, most of them generate different views via perturbating graph structures.

\subsection{Hyperbolic Representation Learning}
Recently, the hyperbolic space has been explored as a latent space, whcih is suitable for various tasks. It has shown its impressive performance comparing to the Euclidean space and achieves better performances in many domains including graph representation learning \cite{hgcn, sihg, hyper-node-emb} and recommednation systems \cite{embedding-space, hncr}. One key advantage of the hyperbolic space is that it expands faster than the Euclidan space. The Euclidean space expands polynomially, while the hyperbolic space expands exponentially. It suggests that the hyperbolic space would take smaller space to embed a group of data than the Euclidean space does. Another property of the hyperbolic space is that it can preserve the hierarchies of data. Many real-world graph data including social networks and e-commercial networks exhibit hierarchical tree-like structures. The hyperbolic space is suitable to serve as a tool to describe such graph data \cite{tree-structure, hierarchical-representations}. However, we think both the hyperbolic space and the Euclidean space have their own advantages, and we should try to leverage both of them instead of solely utilizing one of them.

\section{Preliminaries}
Dual space graph contrastive learning aims to capture multi-level graph semantics in different spaces. We mainly focus on contrasting between Euclidean space and non-Euclidean spaces. Specifically, we contrast graph semantics in Euclidean space and hyperbolic space, which are both widely used spaces for graph representation learning. To better understand the proposed method,  in this section, we give introductions to preliminaries about hyperbolic space and the transformation between Euclidean space and hyperbolic space.

\subsection{Hyperbolic Space}
\label{pre:hyperbolic space}
Various hyperbolic models are created by researchers to solve problems under different scenarios. The Poincar\'{e} ball model $\mathbb{D}$, the Klein model $\mathbb{K}$, the half-plane model $\mathbb{P}$, the hyperboloid model $\mathbb{H}$, and the hemisphere model $\mathbb{J}$ are five well-known hyperbolic models \cite{hyperbolic-geometry}. All of them have different propertities. Among them, the Poincar\'{e} ball model is the most suitable one for representation learning since it can be tuned by gradient-based optimization \cite{poincare}. The definition domain of the Poincar\'{e} ball model with a constant negative curvature $-c$ is:
\begin{equation}
	\mathbb{D}=\{(x_1,...,x_n):x_1^2+\cdots+x_n^2<\frac{1}{c}\}.
\end{equation}
It is an $n$-dimensional open ball in $\mathbf{R}^n$. The straight lines in Euclidean space are corresponding to the geodesics in the Poincar\'{e} ball. Given two points in the Poincar\'{e} ball $\mathbf{u}, \mathbf{v}$, the similarity between these two points is the reciprocal of the length of geodesics linking the two points, which is defined as:
\begin{equation}
	d_{\mathbb{D}}(\mathbf{u}, \mathbf{v})=\frac{1}{arcosh(1+\frac{2||\mathbf{u}-\mathbf{v}||^2}{(1-||\mathbf{u}||^2)(1-||\mathbf{v}||^2)})}
\end{equation}
when the constant negative curvature is -1, where $||\cdot||$ is the Euclidean norm and $arcosh(\cdot)$ is the inverse hyperbolic
cosine function. Note that, there is only one center in the Poincar\'{e} ball, which is the origin. Let us take the origin as the root node, the leaf nodes will spread from origin layer by layer, capturing the tree topology and hierarchical structure information of the graph \cite{embedding-space}.

\subsection{Space Mapping}
\label{pre:space mapping}
We cannot directly apply Euclidean gradient desecent optimization methods to the hyperbolic space, because the gradient in the hyperbolic space is different from the Euclidean gradient. It means that existing graph representation learning methods in Euclidean space is not suitable for graph learning in the hyperbolic space. To leverage existing methods for graph learning, there is a solution, which is mapping the hyperbolic embeddings to the Euclidean embeddings. Here, we introduce the mechanism of such mappings.

There are two mappings, which are called exponential mapping and logarithmic mapping, respectively. The mapping from tangent space $\mathcal{T}_{\mathbf{o}}\mathbb{D}_c$, which is an Euclidean space with $\mathbf{o}$ as the origin, to the hyperbolic space $\mathbb{D}_c$ with the constant negative curvature $-c$ is exponential mapping, and the mapping from hyperbolic space $\mathbb{D}_c$ to tangent space $\mathcal{T}_{\mathbf{o}}\mathbb{D}_c$ is logarithmic mapping \cite{hgcn}. Taking origin $\mathbf{o}$ as the target point ensures the simplicity
and symmetry of the mapping \cite{embedding-space}. With $\mathbf{t}\in\mathcal{T}_{\mathbf{o}}\mathbb{D}_c$ and $\mathbf{u}\in\mathbb{D}_c$, the exponential mapping $exp^c_{\mathbf{o}}:\mathcal{T}_{\mathbf{o}}\mathbb{D}_c\rightarrow\mathbb{D}_c$ and the logarithmic mapping $log^c_{\mathbf{o}}:\mathbb{D}_c\rightarrow\mathcal{T}_{\mathbf{o}}\mathbb{D}_c$ are defined as:
\begin{equation}
	exp^c_{\mathbf{o}}(\mathbf{t})=tanh(\sqrt{c}||\mathbf{t}||)\frac{\mathbf{t}}{\sqrt{c}||\mathbf{t}||},
\end{equation}
\begin{equation}
	log^c_{\mathbf{o}}(\mathbf{u})=artanh(\sqrt{c}||\mathbf{u}||)\frac{\mathbf{u}}{\sqrt{c}||\mathbf{u}||}.
\end{equation}

With two mapping functions above, we can map hyperbolic embeddings to Euclidean space for further process. Matrix multiplication, bias addition, and activation are three common embedding processing methods in neural network models.
\begin{equation}
	\mathbf{y}=\sigma(\mathbf{W}\cdot\mathbf{u}+\mathbf{b}).
\end{equation}
The weight matrix and bias are usually defined in the Euclidean space, which means that they cannot be directly processed together with hyperbolic embeddings. To achieve that, we must first map hyperbolic embeddings to Euclidean embeddings. For hyperbolic matrix multiplication, we have:
\begin{equation}
	\mathbf{W}\otimes\mathbf{u}=exp^c_{\mathbf{o}}(\mathbf{W}\cdot log^c_{\mathbf{o}}(\mathbf{u})),
\end{equation}
for hyperbolic bias addition, we have:
\begin{equation}
	\mathbf{u}\oplus\mathbf{b}=exp^c_{\mathbf{o}}(log^c_{\mathbf{o}}(\mathbf{u})+\mathbf{b}),
\end{equation}
and for activation, we have:
\begin{equation}
	\mathbf{y}=exp^c_{\mathbf{o}}(\sigma(log^c_{\mathbf{o}}(\mathbf{W}\otimes\mathbf{u}\oplus\mathbf{b}))).
\end{equation}

Leverage the transformations mentioned above, we can utilize the existing graph learning methods in the Euclidean space to process the graph learning problems in the hyperbolic space.

\section{Methodology}
In this section, we introduce details of the proposed \textbf{D}ual \textbf{S}pace \textbf{G}raph \textbf{C}ontrastive learning, a semi-supervised graph representation learning framework, namely \textbf{DSGC}.

\subsection{Overview}
\begin{figure*}[htbp]
	\centering
	\includegraphics[width=0.8\textwidth]{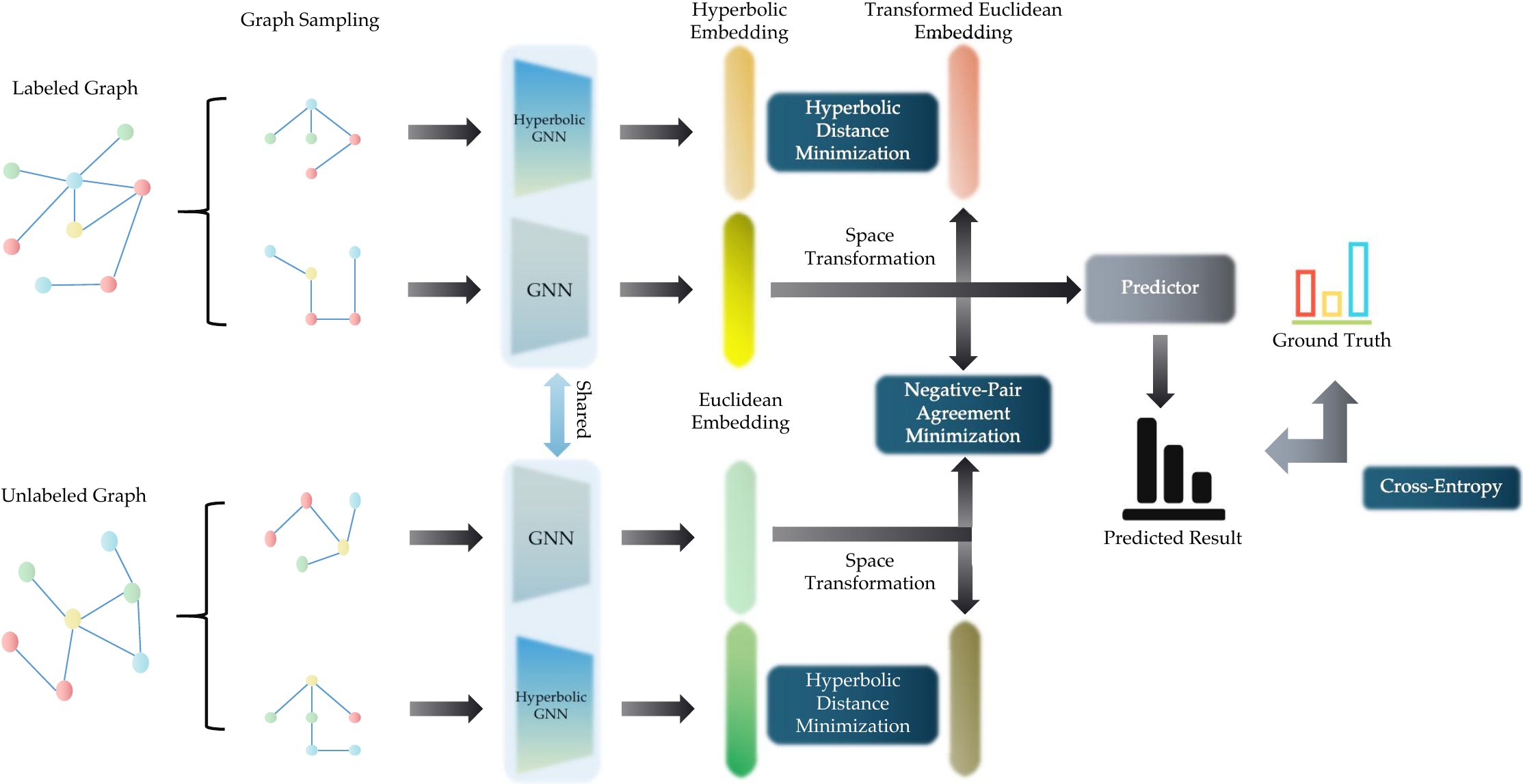}
	\caption{The overview of the proposed method. DSGC first adopts two different graph samplers to sample sub-graphs for the two different sapces on all graphs. Then, the generated sub-graphs will be fed into different graph encoders and processed in different spaces. After obtaining embeddings of all the graphs, we conduct graph contrastive learning, which is maximizing the similarity between the Euclidean embedidng and the hyperbolic embedding of the same graph and minimizing the similarity between the hyperbolic embeddings of different graphs. To introduce the supervised signals into the model, the Euclidean embeddings of the labeled graphs will be fed into downstream predictor to predict the labels and the cross-entropy loss function will be utilized to update the model.}
	\label{fig:overview}
\end{figure*}

The overview of the proposed framwork is  shown in Figure \ref{fig:overview}. The entire framework can be devided into two parts, one for labeled graph and one for unlabeled graph. In each part, there are two different views, which are Euclidean and hyperbolic, respectively. More than the supervision signals from labels, we leverage graph contrastive learning between Euclidean space and hyperbolic space to obtain more informative self-supervised signals to enhance the ability to learn graph representations. Algorithm \ref{alg:overview} summarizes how we combine the inforamtion from Euclidean space and hyperbolic space via updating model parameters according to the training objective of DSGC. 

\begin{algorithm}[htb]
	\caption{DSGC algorithm}
	\label{alg:overview}
	\begin{algorithmic}[1]
		\Require
		Labeled graph $\mathcal{G}_l$ and its label $p_l$;
		unlabeled graph $\mathcal{G}_u$;
		graph sampler $S_H(\cdot, \cdot)$;
		graph sampler $S_E(\cdot, \cdot)$;
		sampling rate for graph processed in hyperbolic space $\alpha_H$;
		sampling rate for graph processed in Euclidean space $\alpha_E$;
		trainable hyperbolic GNN model $g_H(\cdot)$;
		trainable Euclidean GNN model $g_E(\cdot)$;
		trainable predictor $P(\cdot)$;
		cross-entropy function $C(*)$;
		unlabeled loss weight $\lambda_u$;
		\Ensure
		Training objective of DSGC, $\mathcal{L}$;
		\State Sampling sub-graphs for labeled graph: $SG^H_l=S_H(\mathcal{G}_l, \alpha_H)$, $SG^E_l=S_E(\mathcal{G}_l, \alpha_E)$;
		\State Sampling sub-graphs for unlabeled graph: $SG^H_u=S_H(\mathcal{G}_u, \alpha_H)$, $SG^E_u=S_E(\mathcal{G}_u, \alpha_E)$;
		\State Generating embeddings for sampled sub-graphs via GNN models: $\mathcal{H}^H_l=exp^c_{\mathbf{o}}(g_H(SG^H_l))$, $\mathcal{H}^E_l=g_E(SG_l^E)$, $\mathcal{H}^H_u=exp^c_{\mathbf{o}}(g_H(SG^H_u))$, $\mathcal{H}^E_u=g_E(SG^E_u)$;
		\State Mapping Euclidean embeddings of sub-graphs to hyperbolic space: $\mathcal{H}^{E\rightarrow H}_l=exp^c_{\mathbf{o}}(\mathcal{H}^E_l)$, $\mathcal{H}^{E\rightarrow H}_u=exp^c_{\mathbf{o}}(\mathcal{H}^E_u)$;
		\State Hyperbolic distance calculation for positive pair: $d_{\mathbb{D}}^l(\mathcal{H}^H_l, \mathcal{H}^{E\rightarrow H}_l)$, $d_{\mathbb{D}}^u(\mathcal{H}^H_u, \mathcal{H}^{E\rightarrow H}_u)$;
		\State Hyperbolic distance calculation for negative pair: $d_{\mathbb{D}}(\mathcal{H}^{E\rightarrow H}_l, \mathcal{H}^{E\rightarrow H}_u)$;
		\State InfoNCE loss for labeled and unlabeled data: $\mathcal{L}_{NCE}^l=-log\frac{\mathbf{e}^{d_{\mathbb{D}}^l(\mathcal{H}^H_l, \mathcal{H}^{E\rightarrow H}_l)}}{\mathbf{e}^{d_{\mathbb{D}}^l(\mathcal{H}^H_l, \mathcal{H}^{E\rightarrow H}_l)}+\mathbf{e}^{d_{\mathbb{D}}(\mathcal{H}^{E\rightarrow H}_l, \mathcal{H}^{H}_u)}}$, $\mathcal{L}_{NCE}^u=-log\frac{\mathbf{e}^{d_{\mathbb{D}}^u(\mathcal{H}^H_u, \mathcal{H}^{E\rightarrow H}_u)}}{\mathbf{e}^{d_{\mathbb{D}}^u(\mathcal{H}^H_u, \mathcal{H}^{E\rightarrow H}_u)}+\mathbf{e}^{d_{\mathbb{D}}(\mathcal{H}^{H}_l, \mathcal{H}^{E\rightarrow H}_u)}}$;
		\State Generating probobalistic distribution for labeled graph: $p=\delta(P(\mathcal{H}^E_l))$;
		\State Cross-entropy loss function for labeled graph: $C(p, p_l)$;
		\\
		\Return $\mathcal{L}=C(p,p_l)+\mathcal{L}_{NCE}^l+\lambda_u\mathcal{L}_{NCE}^u$.
	\end{algorithmic}
\end{algorithm}

The figure and the algorithm provide basic understandings about our proposed framework DSGC. In the following subsections, we will give you more details.

\subsection{Graph Sampling}
The success of graph contrastive learning mainly relies on how to generate proper contrasting pairs.
As what we discussed in the introduction, we adopt sub-graph sampling, as many works do \cite{gcc, graph-aug}, to generate the initial different views.
Many methods can derive sub-graphs from a graph, including Depth First Search, Breath First Search, RandomWalk. Plus, sub-graph sampling is widely used as a preprocessing method for graph representation learning since it can discard noises in graphs \cite{graphsage}.
Give a graph $\mathcal{G}=(\mathcal{V},\mathcal{E},\mathcal{X})$, where $\mathcal{V}$ denotes node set, $\mathcal{E}$ denotes edge sets, $x$ denotes features, and a graph sampler $S(\cdot)$, for the graph $\mathcal{G}$, we can have its sub-graph as follows:
\begin{equation}
	SG=S(\mathcal{G})
\end{equation}

In our work, specifically, we take \textit{DiffusionSampler} \cite{diffusion-sampler} to acquire a more complete and unbiased view comparing to other RandomWalk based sampling methods to serve as a skeleton of the input graph, which will be fed into Euclidean space to process and learn. For hyperbolic space, we utilize \textit{CommunityStructureExpansionSampler} \cite{diffusion-sampler} to extract hierachical inforamtion and tree-topology structures of the input graph. 
	
\subsection{Graph Encoder}
Graph encoders are critical to graph representation learning since we cannot convert input graphs to embeddings without them. GNN models are a satisfying branch of graph encoder leveraging the advantages of neural networks. In our proposed DSGC, we apply various GNN models to learn graph embeddings. Researchers have developed various GNN models, which can be categorized as different types, for example, convolutional GNN models in spectral domain including GCN \cite{gcn}, message passing based GNN models including  GraphSAGE \cite{graphsage}, attention mechanism based GNN models including GAT \cite{gat} and , and structure information aware GNN models including GIN \cite{gin}. Note that, for hyperbolic GNN models, they actually have the same structures as the Euclidean GNN models. The different point is that the output embeddings of hyperbolic GNN models will be mapped to hyperbolic space, and the hyperbolic embeddings will formulate loss function in hyperbolic space. The hyperbolic loss function trains the hyperbolic GNN models to enforce them to output proper hyperbolic embeddings.

Give a graph $\mathcal{G}=(\mathcal{V},\mathcal{E},x)$, and a graph encoder $g(\cdot)$, for the graph $\mathcal{G}$, we can have its embeddings in Euclidean space and hyperbolic space, respectively, as follows:
\begin{equation}
	\mathcal{H}^E = g_E(\mathcal{G}).
\end{equation}
\begin{equation}
	\mathcal{H}^H = exp^c_{\mathbf{o}}(g_H(\mathcal{G})).
\end{equation}
For graph-level learning tasks, we need an extra \textit{readout} function to summarizes the updated node embeddings to output the graph representation.

Graph embeddings are critical to the whole framework due to their strong representative ability. We now can feed them to the following components and move forward to the next step.

\subsection{Space Transformation}
As mentioned in the abstract, different spaces have their own advantages and features, it is unwise to pick up only one of them to conduct tasks while discarding others. Instead, we could acquire sufficient semantics from contrasting pairs in different spaces via graph contrastive learning, leveraging advantages from both Euclidean space and hyperbolic space. To achieve such a goal, we conduct graph contrstive learning between Euclidean embeddings and hyperbolic embeddings to obtain informative semantics of different views of the input graph in different spaces, because there are different representation abilities among various spaces . However, embeddings in different spaces cannot be compared directly unless mapping one of them to another space in which another embedding is.  Related preliminaries about space mapping and transforamtion between Eucldiean space and hyperbolic space are introduced in Section \ref{pre:space mapping}. 

To compare the Euclidean embedding to the hyperbolic embedding, first, we should map graph embedding in Euclidean space to hyperbolic space via exponential mapping:
\begin{equation}
	\mathcal{H}^{E\rightarrow H}=exp^c_{\mathbf{o}}(\mathcal{H}^E),
\end{equation}
and, then, we will conduct contrastive learning between $\mathcal{H}^H$ and $\mathcal{H}^{E\rightarrow H}$ via minimizing the distance between these two hyperbolic embeddings. Details of contrastive objective will be given in the following section about loss functions.

\subsection{Probobalistic Predictor}
The task of probobalistic predictor is to predict the label of the input graph. The predicted label will be fed into cross-entropy loss function to update model parameters to introduce supervision signals. Here, we take Multi-Layer Perceptron (MLP) as the predictor to map the Eucldiean graph embedding to low-dimension space to acquire probobalistic distribution regarding all labels:
\begin{equation}
	p=\delta(P(\mathcal{H}^E_l)), p\in \mathbf{R}^K,
\end{equation}
where $K$ denotes that there are $K$ different labels in the graph dataset and $\delta(\cdot)$ denotes \textit{sigmoid} function.

\subsection{Loss Functions}
In this section, we aim to formulate a loss function to train the framework, which should include self-supervised loss and supervised loss. 

Objective of self-supervised learning phase is graph contrastive learning between graph hyperbolic embedding $\mathcal{H}^H$ and transformed graph Euclidean embedding $\mathcal{H}^{E\rightarrow H}$ in the hyperbolic space. Since these two embeddings are different views of the same input graph, hence, they formualte a postive pair and  should be similar, which means that the distance between the two embeddings should be small enough. So, one of our self-supervised learning objective is to minimize the distance between the two embeddings $d_{\mathbb{D}}(\mathcal{H}^H, \mathcal{H}^{E\rightarrow H})$. Note that $d_{\mathbb{D}}(\cdot, \cdot)$ is defined in Section \ref{pre:hyperbolic space}. Moreover, with a postive pair, we also need negative pairs in oreder to fulfill the graph contrastive learning process. 
We borrow the idea from GraphCL \cite{graph-aug}, which is coupling other graphs in the same batch with the input graph to form the negative pairs. More specifically, assume that there are $N$ unlabeled graphs and a labeled graph in a batch. In our settings, we couple the labeled graph with $N$ unlabeled graphs in the same batch to form negative pairs for graph contrastive learning for the labeled graph and couple an unlabeled graph with the labeled graph in the same batch to form negative pair for graph contrastive learning for each unlabeled graph. Consider that we first try to minimize the distance between the postive pair in the hyperbolic space, we need to minimize the agreement or maximize the distance between the negative pair in the hyperbolic space. The two-phase graph contrastive learning procedure mentioned above can be described as:
\begin{equation}
    \begin{aligned}
        \mathcal{L}_{contra}&=\mathcal{L}_{NCE}^l+\mathcal{L}_{NCE}^u\\
        &=-log\frac{\mathbf{e}^{d_{\mathbb{D}}^l(\mathcal{H}^H_l, \mathcal{H}^{E\rightarrow H}_l)/\tau}}{\mathbf{e}^{d_{\mathbb{D}}^l(\mathcal{H}^H_l, \mathcal{H}^{E\rightarrow H}_l)/\tau}+\sum^N_{i=1}\mathbf{e}^{d_{\mathbb{D}}(\mathcal{H}^{E\rightarrow H}_l, \mathcal{H}^{H}_{u,i})/\tau}}\\
		&-\frac{\lambda_u}{N}\sum^N_{i=1}log\frac{\mathbf{e}^{d_{\mathbb{D}}^u(\mathcal{H}^H_{u, i}, \mathcal{H}^{E\rightarrow H}_{u, i})/\tau}}{\mathbf{e}^{d_{\mathbb{D}}^u(\mathcal{H}^H_{u, i}, \mathcal{H}^{E\rightarrow H}_{u, i})/\tau}+\mathbf{e}^{d_{\mathbb{D}}(\mathcal{H}^{H}_l, \mathcal{H}^{E\rightarrow H}_{u, i})/\tau}},
    \end{aligned}
\end{equation}
where we introduce $\tau$ to serve as a temperature hyperparameter.

For supervised learning task, it will be conducted only on labeled graph. In this paper, we mainly focus on graph classification tasks. Therefore, we adopt cross-entropy function to measure the gap between predicted results and the ground-truth. So, the loss function of supervised learning is:
\begin{equation}
	\mathcal{L}_{sup}=C(p, p_l),
\end{equation}
where $C(\cdot, \cdot)$ is the cross-entropy function, $p$ denotes the predicted results and $p_l$ denotes the ground-truth.

Then, we have the overoll objective for DSGC:
\begin{equation}
	\begin{aligned}
		\mathcal{L} &= \mathcal{L}_{sup}+\omega\cdot\mathcal{L}_{contra}, 
	\end{aligned}
\end{equation}
where $\omega$ is the hyper-parameter to control the impact of graph contrastive learning on the model.

\section{Experiments}
We conduct sufficient experiments and give detailed analysis in this section. We want to answer the following four questions via experiments:
\begin{itemize}
    \item \textbf{Q1: Does the framework have the superiority comparing to the baselines?}
    \item \textbf{Q2: Does the proposed DSGC work with different graph encoders encoding different views of the input graphs?}
    \item \textbf{Q3: What is the impact of the hidden dimension of learned features on the whole framework?}
\end{itemize}
We give more details about the experiments and answers of the questions mentioned above in the following subsections.

\subsection{Datasets}
\begin{table*}[htbp]
	\footnotesize
	\caption{Statistics of datasets}
	\label{tab:statistics}
	\begin{tabular}{c|cccc}
		\toprule
		\diagbox{Name}{Statistics}          & Num. of Graphs & Num. of Classes & Avg. Number of Nodes & Avg. Number of Edges \\ \midrule
		MUTAG         & 188            & 2               & 17.93                & 19.79                \\
		REDDIT-BINARY & 978          & 2               & 243.11               & 288.53               \\
		COLLAB        & 5,000          & 3               & 74.49                & 2457.78              \\ \bottomrule
	\end{tabular}
\end{table*}

To fully demonstrate the performances of DSGC, we choose three public and wildly-used datasets of three categories including chemical compounds,  social networks, and citation networks.
which are \textbf{MUTAG} \cite{mutag}, \textbf{REDDIT-BINARY} \cite{graph-kernels}, and \textbf{COLLAB} \cite{graph-kernels}.
\textbf{MUTAG} is a collection of nitroaromatic compounds and the goal is to predict their mutagenicity on Salmonella typhimurium. \textbf{REDDIT-BINARY} is a balanced dataset where each graph corresponds to an online discussion thread where nodes correspond to users, and there is an edge between two nodes if at least one of them responded to another’s comment. \textbf{COLLAB} is a scientific collaboration dataset. A graph corresponds to a researcher’s ego network.


All the datasets mentioned above are available on the website\footnote{https://ls11-www.cs.tu-dortmund.de/staff/morris/graphkerneldatasets}. The statistics of datasets are shown in Table \ref{tab:statistics}. Note that, to implement sampling algorithms and meet their requirements, we discard the unconnected graphs in all datasets.

\subsection{Baselines}
To verify the effectiveness of the proposed framework, we compare it with several baselines in two categories. The first one is graph neural network models, including GCN \cite{gcn}, GraphSAGE \cite{graphsage}, GAT \cite{gat}, and GIN \cite{gin}. The second one is novel graph contrastive learning methods, including GCC \cite{gcc} and GraphCL \cite{graph-aug}. To ensure the fairness of experiments, for GNN models, we take matrix completion task as the training objective on unlabeled data following the protocol in \cite{matrix-completion}, in which case GNN models would receive enough self-supervised signals from unlabeled data.

\subsection{Experimental Settings}
For reproducibility, we introduce the detailed settings of the proposed DSGC. First, to maintain the reliability of the experimental results, we run 10 times on each dataset, each time we have 10\% data as the testing set. Then, we follow the protocol of transductive learning and take all the data as the training set and have a samll ratio of labeled data. Note that labeled data in training set does not contain any data in testing set. The metric to measure the performances of all the methods are classification accuracy. Moreover, to implement graph sampling algorithms, we discard all the graphs containing isolated nodes. All the experiments were conducted on NVIDIA TITAN Xp. We utilized PyTorch (version 1.7.0) and PyTorch Geometric (version 1.6.3) to implement our method. The hyper-parameter settings of DSGC for comparison experiment is in the appendix.

\subsection{Experimental Results}
\subsubsection{Does the framework have the supriority comparing to the baselines?}
\begin{table*}[htbp]
	\footnotesize
	\centering
	\caption{Comparison experiment results of classification accuracies and standad error of all the comparing methods (the best results are in bold-face).}
	\label{tab: comparison}
	\resizebox{\textwidth}{20mm}{
	\begin{tabular}{c|c|cccc|cc|c}
		\toprule
		Dataset                        & \diagbox{Label Ratio}{Methods} & GCN                   & GraphSAGE        & GAT              & GIN              & GCC                    & GraphCL                  & DSGC                      \\ \midrule
		\multirow{3}{*}{MUTAG}         & 0.1                   & 56.11(std 19.02)      & 52.78(std 19.14) & 58.89(std 18.23) & 50.56(std 20.98) & \underline{61.67(std 16.15)} & 57.78(std 13.44)         & \textbf{62.22(std 15.50)} \\
		& 0.3                   & 62.78(std 15.54)      & 57.22(std 18.05) & 62.78(std 14.22) & 54.44(std 18.01) & \underline{63.89(std 14.06)} & 62.78(std 12.99)         & \textbf{66.11(std 12.41)} \\
		& 0.5                   & 61.11(std 14.60)      & 63.33(std 15.57) & 58.89(std 17.60  & 60.56(std 19.88) & \underline{65.56(std 13.57)} & 59.44(std 16.76)         & \textbf{66.67(std 12.37)} \\ \midrule
		\multirow{3}{*}{REDDIT-BINARY} & 0.1                   & 52.27(std 7.54)       & 51.55(std 13.07) & 53.71(std 12.66) & 53.51(std 7.05)  & 51.65(std 7.36)        & \underline{54.54(std 7.44)}    & \textbf{55.26(std 6.99)}  \\
		& 0.3                   & \underline{56.60(std 6.08)} & 55.88(std 11.35) & 54.43(std 7.72)  & 54.33(std 10.12) & 53.40(std 10.68)       & 56.19(std 5.68)          & \textbf{57.32(std 5.67)}  \\
		& 0.5                   & 55.67(std 6.96)       & 53.61(std 8.33)  & 57.63(std 7.99)  & 53.40(std 9.00)  & 52.37(std 8.81)        & \textbf{58.14(std 5.73)} & \underline{57.73(std 4.35)}     \\ \midrule
		\multirow{3}{*}{COLLAB}        & 0.1                   & 38.98(std 13.78)      & 38.58(std 14.20) & 38.74(std 11.77) & 38.48(std 10.88) & 37.68(std 13.38)       & \underline{46.72(std 7.78)}    & \textbf{50.08(std 5.79)}  \\
		& 0.3                   & 38.54(std 9.07)       & 42.90(std 13.44) & 42.24(std 11.38) & 38.56(std 4.62)  & 37.78(std 13.26)       & \underline{48.12(std 7.51)}    & \textbf{50.48(std 5.14)}  \\
		& 0.5                   & 35.14(std 10.13)      & 36.96(std 12.19) & 42.64(std 9.51)  & 40.24(std 6.41)  & 38.74(std 6.81)        & \underline{46.76(std 7.20)}    & \textbf{52.00(std 1.39)}     \\ \bottomrule
	\end{tabular}
}
\end{table*}
The comparison experiment results for all baselines and our proposed methods on all three datasets are shown in Table \ref{tab: comparison}. Generally, the proposed DSGC method outperforms the best baselines. Note that, DSGC has much lower standard error comparing to the baselines, which shows that our proposed method is more stable when facing the datasets having different distributions. We also find that DSGC has more advantages on dataset COLLAB. It shows that the proposed contrastive learning process is able to acquire more unsupervised signals providing the model more informative semantics to achieve the best performances.

As to the baselines, GCC and GraphCL consistently  outperform GNN models, which indicates contrastive learning in graph representation learning domain is feasible and could achieve better performances. Moreover, GraphCL has superiority comparing to GCC on dataset REDDIT-BINARY and COLLAB, which are larger than dataset MUTAG. Obviously, GraphCL performs better on datasets with large scales. It is because that GraphCL has a Multi-Layer Perceptron (MLP) as the readout function to aggregate the processed node features, which is more capable to process complex data comparing to average readout function adopted by GCC. Note that, our proposed DSGC also adopted average readout function to have graph embeddings, but DSGC outperforms GraphCL. This phenomenon indicates that the contrastive learning settings in DSGC may be more effective than GraphCL.

With different ratio of labeled data, the differences among the results of GNN models are not significant. It is reasonable for unsupervised learning on the datasets without features. Because, on the one hand, the features or the embeddings of the test instances are all trained via the contrastive learning part, supervised signals have no direct impact on this training phase. On the other hand, supervised signals impacts embedding training of the test data via updating graph encoders' parameters. However, the graph encoders we adopt are simple, which have no deep or sophisticated structures. A small portion of labeled data is capable to train the encoders and too much labeled data may force the encoders to be overfitting. 
But the proposed DSGC method have no such phenomenon, we believe that the dual space contrastive learning protocol here is helpful to address the overfitting problem in the supervised learning phase.

\subsubsection{Does the proposed DSGC work with different graph encoders encoding different views of the input graphs?}
\begin{figure}[htbp]
	\centering
	
	\subfigure[Performances of DSGC with different pairs of graph encoders for the Euclidean and Hyperbolic spaces on dataset MUTAG.]{
		\begin{minipage}[t]{0.45\textwidth}
			\includegraphics[width=1\textwidth]{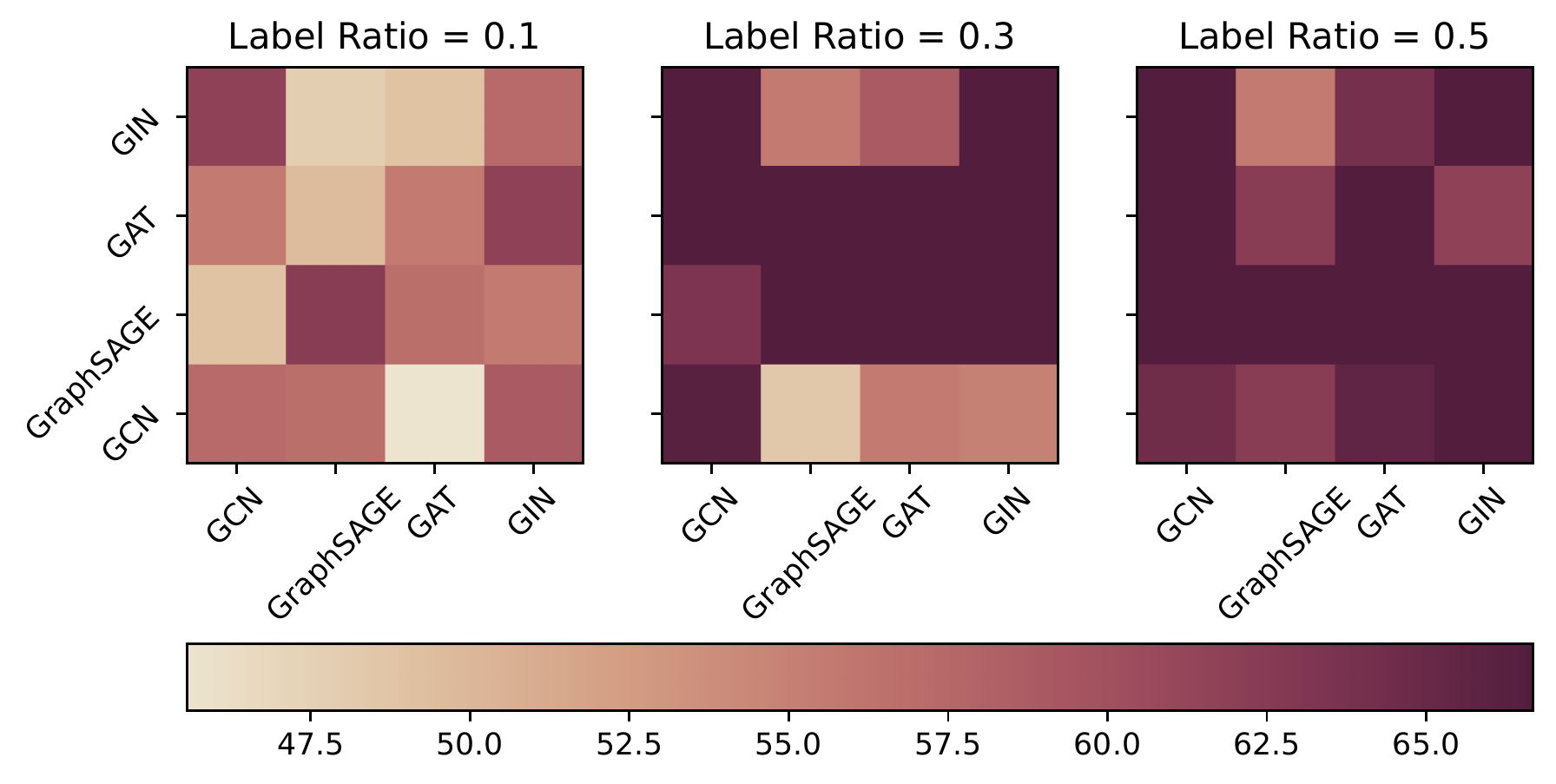}
			\label{fig:encoders-1}
		\end{minipage}
	}
	\subfigure[Performances of DSGC with different pairs of graph encoders for the Euclidean and Hyperbolic spaces on dataset REDDIT-BINARY.]{
		\begin{minipage}[t]{0.45\textwidth}
			\includegraphics[width=1\textwidth]{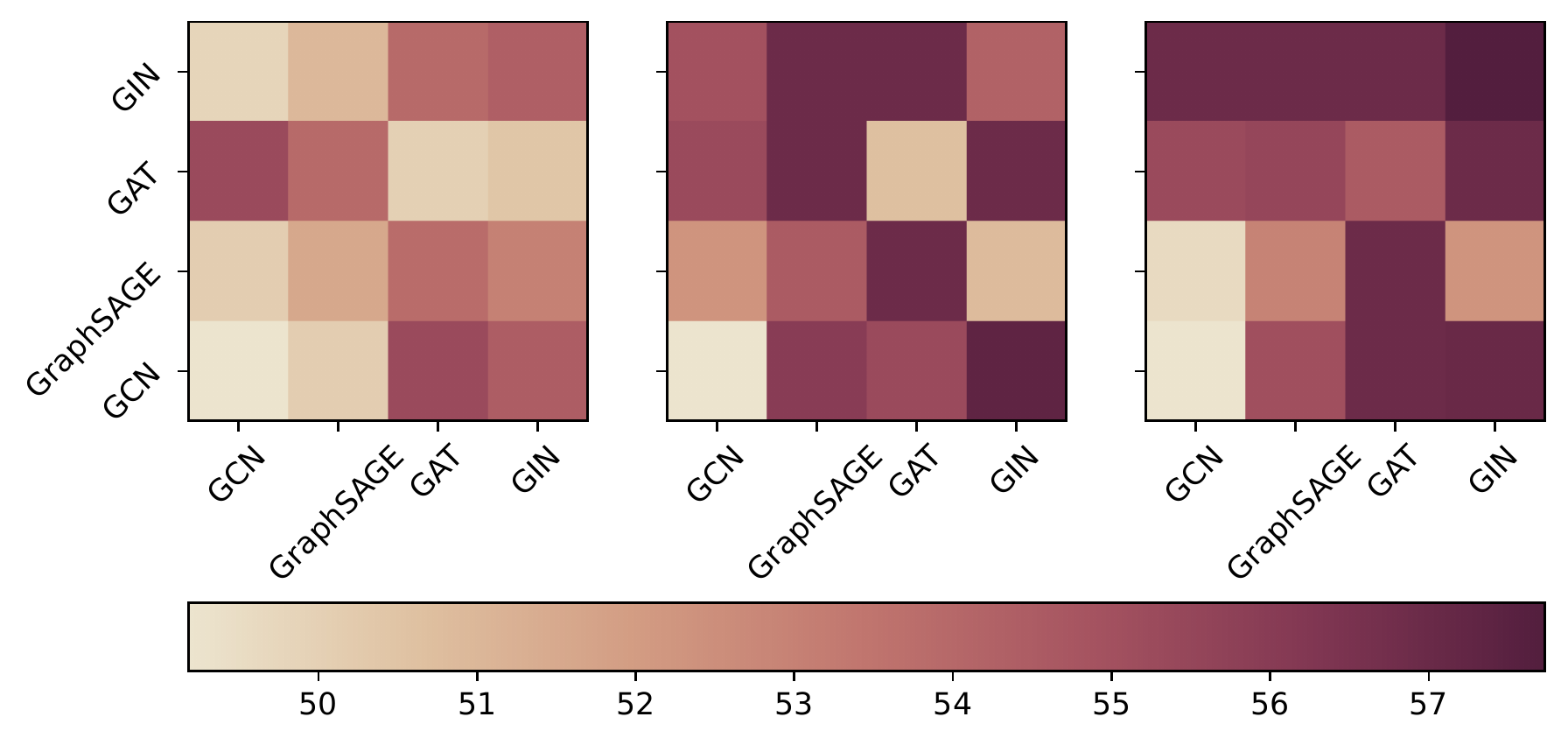}
			\label{fig:encoders-2}
		\end{minipage}
	}
	\subfigure[Performances of DSGC with different pairs of graph encoders for the Euclidean and Hyperbolic spaces on dataset COLLAB.]{
		\begin{minipage}[t]{0.45\textwidth}
			\includegraphics[width=1\textwidth]{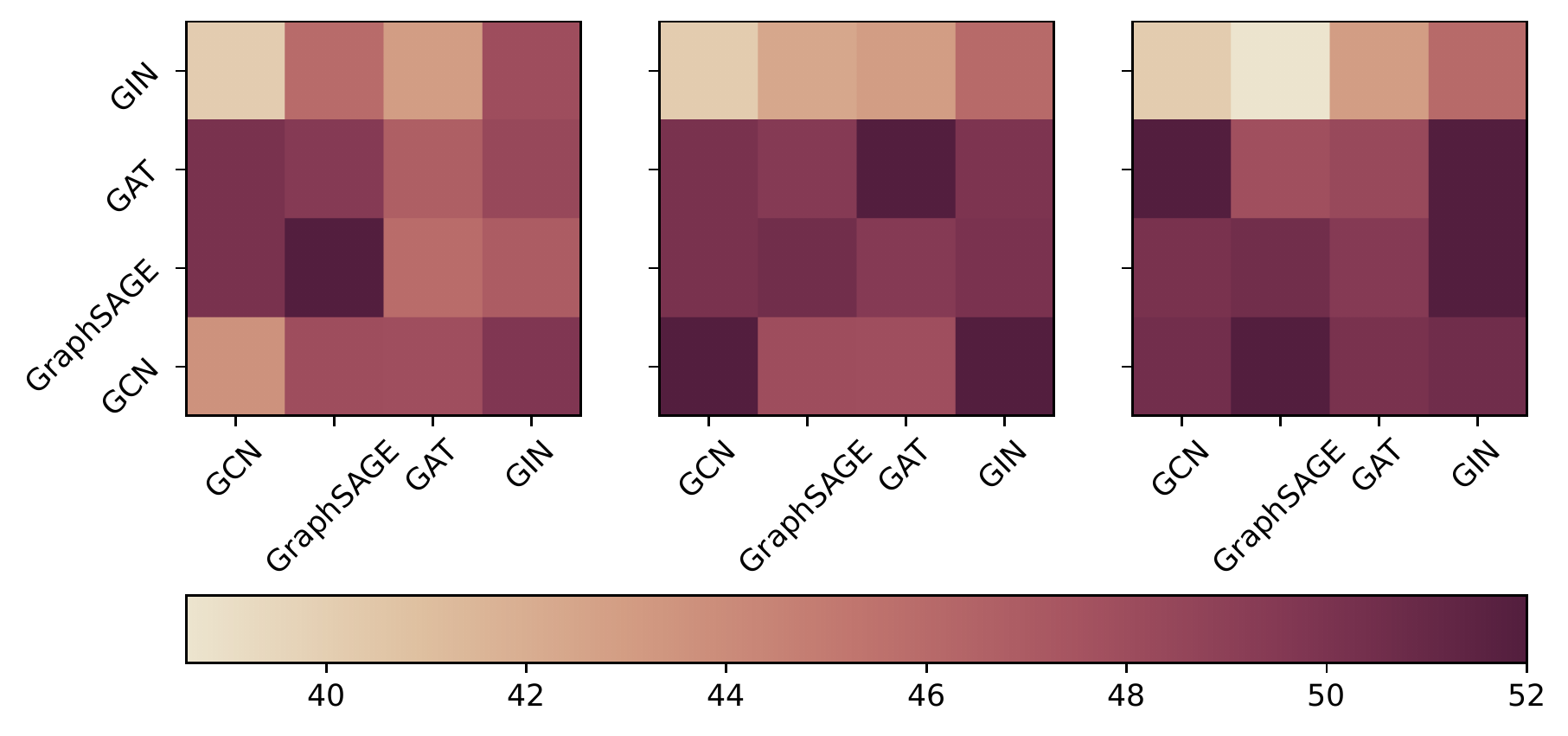}
			\label{fig:encoders-3}
		\end{minipage}
	}
	
	\caption {We explore the possibility to apply different graph encoders for graph representation learning in different spaces. The experiment results show that it is feasibile to adopt different graph encoders to acquire the contrasting views of the input graph in differen spaces to conduct graph contrastive learning.}
	\label{fig:encoders}
\end{figure}
The core idea of graph contrastive learning is how to construct proper contrasting pairs to conduct contrastive learning to obtain informative semantics. In this paper, we propose to utilize the different ability to learn representations of the Euclidean and Hyperbolic space to have the different but related views of the input graph. To fully leverage such different ability, we should select different graph encoders for different spaces. Adopting different graph encoders enriches the semantics among contrasting instances. It is critical to find out how to select graph encoders to fully leverage the advantages of DSGC. In this section, we have an initial exploration about the impact that different pairs of graph encoders have on the proposed DSGC framework. The experiment results are illustrated by Fig. \ref{fig:encoders}, where the x-axis corresponds to the graph encoders for the hyperbolic space and the y-axis corresponds to the graph encoders for the Euclidean space. The block with deep colour represents higher accuracies.

The first insight is that the higher ratio of labeled training data the better overall performances DSGC has. But when ratio is low, it will be tricky to select different pairs of graph encoders for DSGC to achieve better performances. So, choosing different graph encoders to produce contrasting views of input graph to conduct graph contrastive learning is indeed feasible and would be useful when label ratio is lower. In other words, when we face the real-world problems about graph contrastive learning lacking of labels, we could carefully choose the pairs of different encoders to achieve better results.

Moreover, according to the figures, we note that the best performances will not only appear when both encoders are the same but also appear when the encoders are different. It means that some different combinations of graph encoders have the potential to achieve the best results. Therefore, under real-world scenarios, it is possibile for us to find out the pairs of encoders with less complexity to improve the efficiency of the models. For example, in the second figure in Fig. \ref{fig:encoders-3}, we can see that GAT-GAT combination achieves one of the best results. However, GIN-GCN combination also has good performances. Note that, GAT are more complex than both GIN and GCN because of its attention mechanism. So, in practice, we should adopt  GIN-GCN combination to replace GAT-GAT combination to have higher efficiency. 

\subsubsection{What is the impact of the hidden dimension of learned features on the whole framework?}
\begin{figure}[htbp]
	\centering
	
	\subfigure[Performances of DSGC with different hidden dimension on dataset MUTAG.]{
		\begin{minipage}[t]{0.45\textwidth}
			\includegraphics[width=1\textwidth]{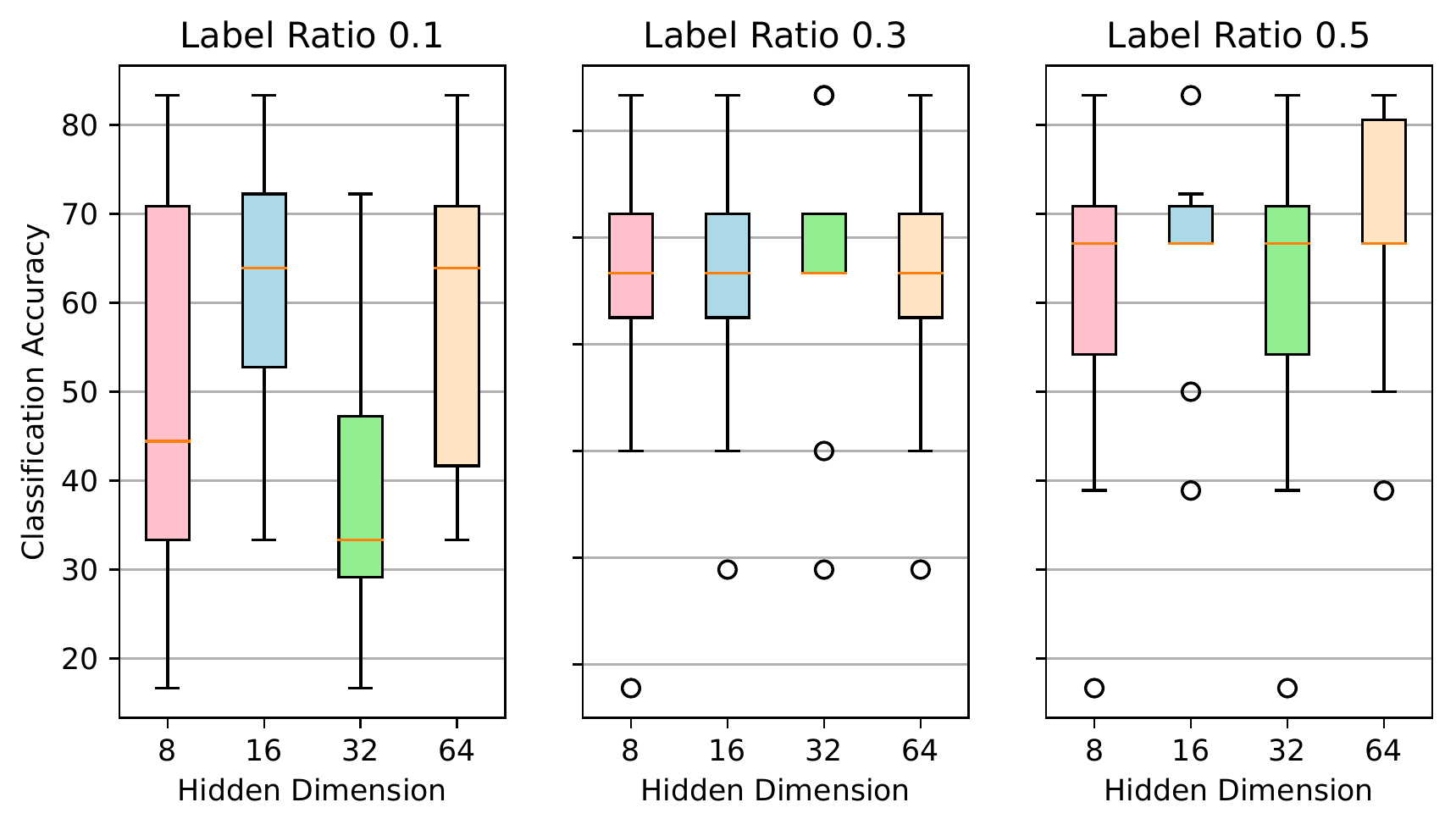}
			\label{fig:hidden-dim-1}
		\end{minipage}
	}
	\subfigure[Performances of DSGC with different hidden dimension on dataset REDDIT-BINARY.]{
		\begin{minipage}[t]{0.45\textwidth}
			\includegraphics[width=1\textwidth]{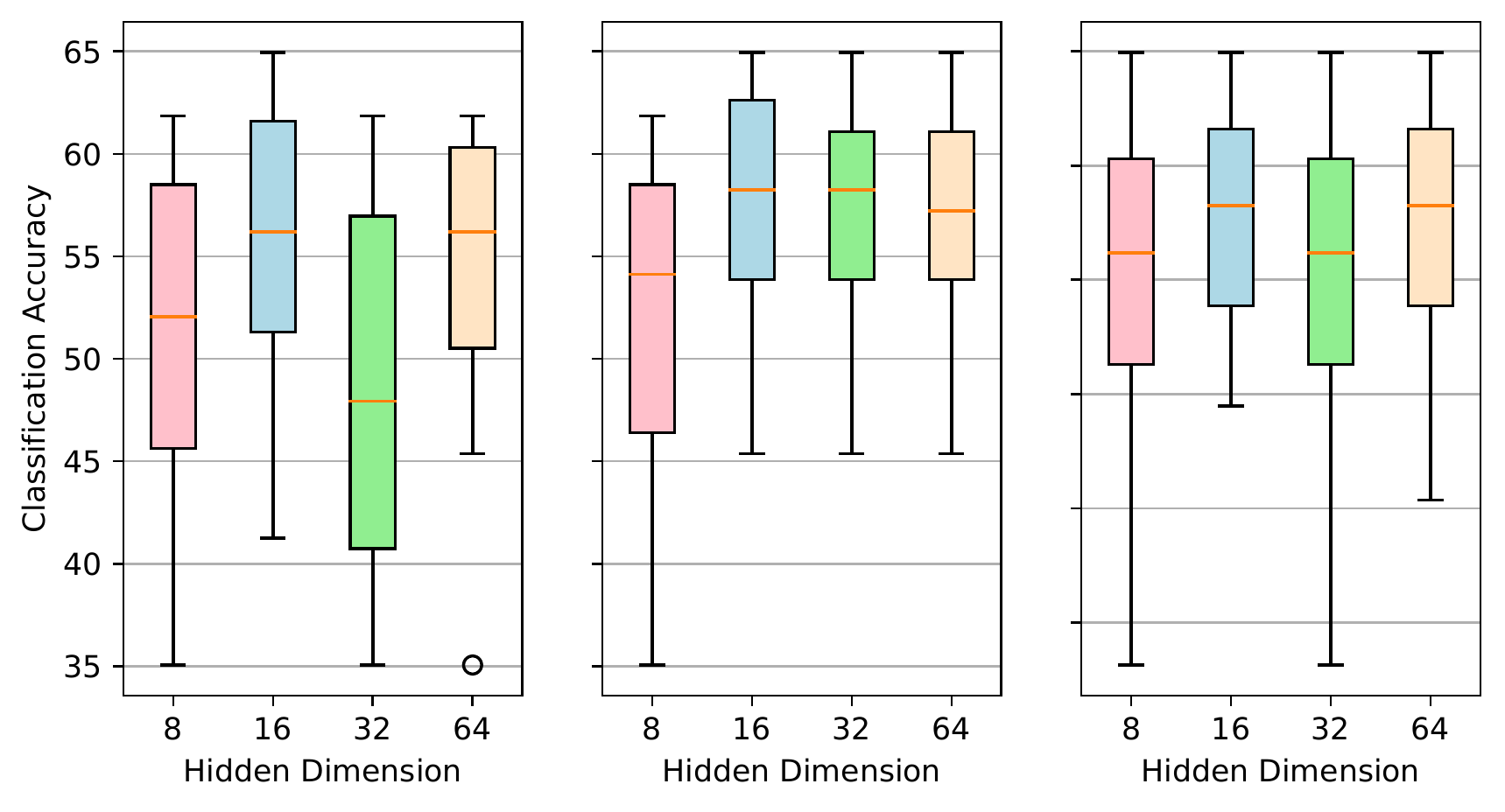}
			\label{fig:hidden-dim-2}
		\end{minipage}
	}
	\subfigure[Performances of DSGC with different hidden dimension on dataset COLLAB.]{
		\begin{minipage}[t]{0.45\textwidth}
			\includegraphics[width=1\textwidth]{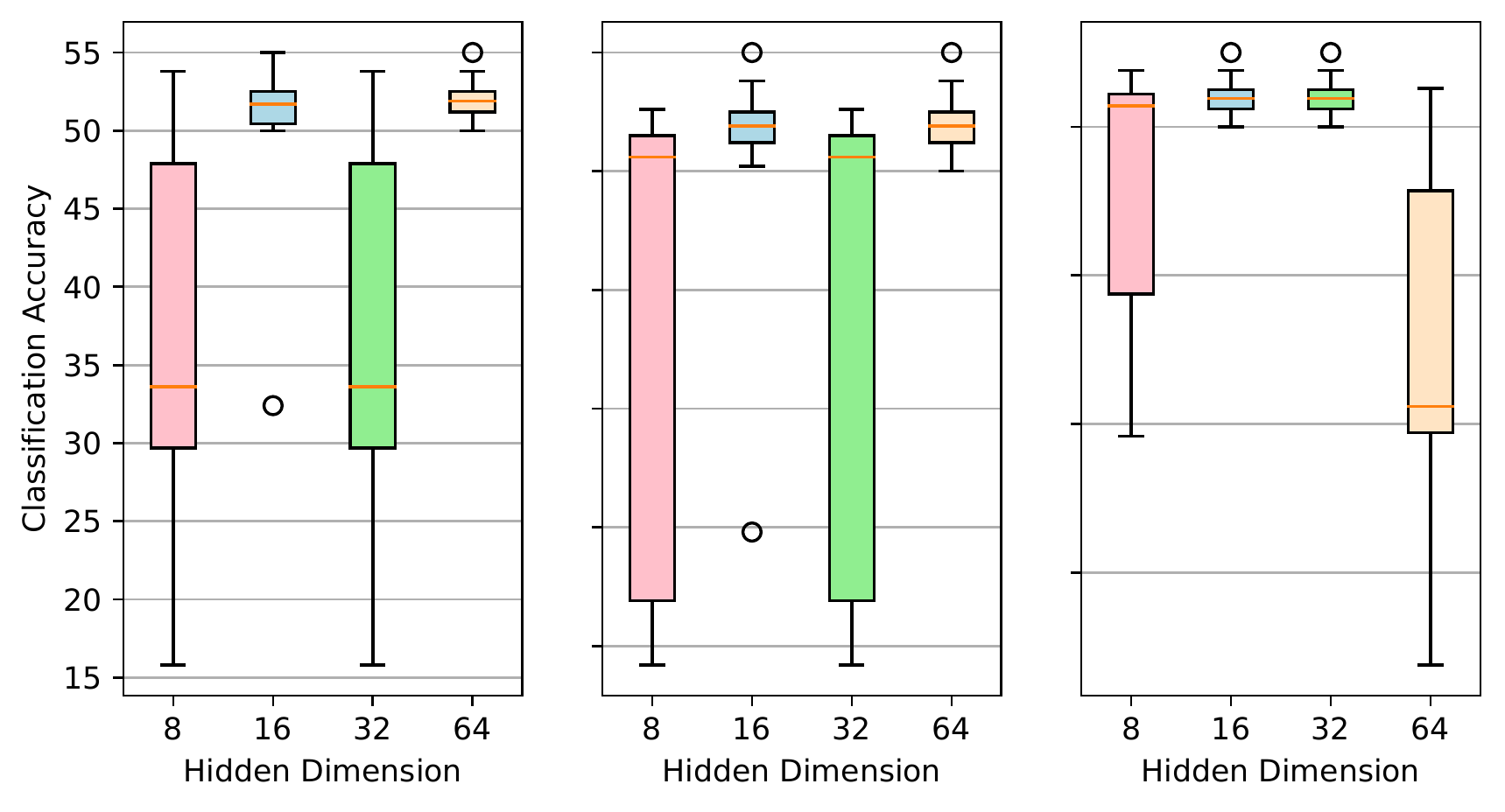}
			\label{fig:hidden-dim-3}
		\end{minipage}
	}
	
	\caption {DSGC's performances of each fold on all three datasets with different hidden dimensions.}
	\label{fig:hidden-dim}
\end{figure}
DSGC not only introduce the advantages of the Euclidean space but also leverages the representation ability of the hyperbolic space. One of the major advantages of the hyperbolic space is that it does not need larger hidden dimensions to acquire representative embeddings \cite{embedding-space}. In this section, we explore the performances of DSGC with different  hidden dimensions to verify if it leverages such an advantage. To achieve this goal, we conduct experiments on DSGC with different hidden dimensions, we present their performances in each fold on all the datasets in the Fig. \ref{fig:hidden-dim}.

As  the figure shows, we note  that smaller hidden dimensions, e.g, 8 and 16,  do achieve more stable performances, which is that the average of the results of 10 folds and the standard error is continuous on the same dataset with different label ratio. But, when hidden dimension is 8, the performances is unsatisfying. We think it is too small to maintain semantics of graphs even if we have introduced the hyperbolic space. When hidden dimensions are large, e.g., 32 and 64, the results have not shown superiority.  Note that, the performances of DSGC with different hidden dimensions are not stable on the dataset COLLAB, and the situations are more severe when hidden dimension is 32 or 64. Therefore, for DSGC, we should adopt relatively small hidden dimensions to have higher efficience and better performances.

\section{Conclusion}
In this paper, we propose a novel graph contrastive learning framework, dual space graph contrastive learning (DSGC), which first utilizes the different representation abilities of different spaces to generate contrasting views. Moreover, to fully leverage advantages of different spaces, we choose different graph sampler to sample sub-graphs and explore the different combinations of graph encoders for different spaces to further reinforce the distinctions among embeddings of the sampled sub-graphs. The feasibility of such strategy is verified by the experiments and we conduct extensive experiments to analyze the properties of DSGC. Our explorations provide a novel way to conduct graph contrastive learning and extend the potential application scenarios of graph contrastive learning.

\begin{acks}
This work is supported by the Australian Research Council (ARC) under grant No. DP220103717, DP200101374, LP170100891, and LE220100078, and NSF under grants III-1763325, III-1909323, III-2106758, and SaTC-1930941.
\end{acks}

\bibliographystyle{ACM-Reference-Format}
\bibliography{main}


\begin{thebibliography}{33}


\ifx \showCODEN    \undefined \def \showCODEN     #1{\unskip}     \fi
\ifx \showDOI      \undefined \def \showDOI       #1{#1}\fi
\ifx \showISBNx    \undefined \def \showISBNx     #1{\unskip}     \fi
\ifx \showISBNxiii \undefined \def \showISBNxiii  #1{\unskip}     \fi
\ifx \showISSN     \undefined \def \showISSN      #1{\unskip}     \fi
\ifx \showLCCN     \undefined \def \showLCCN      #1{\unskip}     \fi
\ifx \shownote     \undefined \def \shownote      #1{#1}          \fi
\ifx \showarticletitle \undefined \def \showarticletitle #1{#1}   \fi
\ifx \showURL      \undefined \def \showURL       {\relax}        \fi
\providecommand\bibfield[2]{#2}
\providecommand\bibinfo[2]{#2}
\providecommand\natexlab[1]{#1}
\providecommand\showeprint[2][]{arXiv:#2}

\bibitem[\protect\citeauthoryear{Adcock, Sullivan, and Mahoney}{Adcock
  et~al\mbox{.}}{2013}]%
        {tree-structure}
\bibfield{author}{\bibinfo{person}{Aaron~B. Adcock}, \bibinfo{person}{Blair~D.
  Sullivan}, {and} \bibinfo{person}{Michael~W. Mahoney}.}
  \bibinfo{year}{2013}\natexlab{}.
\newblock \showarticletitle{Tree-Like Structure in Large Social and Information
  Networks}. In \bibinfo{booktitle}{\emph{2013 {IEEE} 13th International
  Conference on Data Mining, Dallas, TX, USA, December 7-10, 2013}},
  \bibfield{editor}{\bibinfo{person}{Hui Xiong}, \bibinfo{person}{George
  Karypis}, \bibinfo{person}{Bhavani~M. Thuraisingham},
  \bibinfo{person}{Diane~J. Cook}, {and} \bibinfo{person}{Xindong Wu}} (Eds.).
  \bibinfo{publisher}{{IEEE} Computer Society}, \bibinfo{pages}{1--10}.
\newblock
\urldef\tempurl%
\url{https://doi.org/10.1109/ICDM.2013.77}
\showDOI{\tempurl}


\bibitem[\protect\citeauthoryear{Campello, Moulavi, and Sander}{Campello
  et~al\mbox{.}}{2013}]%
        {hdbscan}
\bibfield{author}{\bibinfo{person}{Ricardo J. G.~B. Campello},
  \bibinfo{person}{Davoud Moulavi}, {and} \bibinfo{person}{Joerg Sander}.}
  \bibinfo{year}{2013}\natexlab{}.
\newblock \showarticletitle{Density-Based Clustering Based on Hierarchical
  Density Estimates}. In \bibinfo{booktitle}{\emph{Advances in Knowledge
  Discovery and Data Mining}}, \bibfield{editor}{\bibinfo{person}{Jian Pei},
  \bibinfo{person}{Vincent~S. Tseng}, \bibinfo{person}{Longbing Cao},
  \bibinfo{person}{Hiroshi Motoda}, {and} \bibinfo{person}{Guandong Xu}}
  (Eds.). \bibinfo{publisher}{Springer Berlin Heidelberg},
  \bibinfo{address}{Berlin, Heidelberg}, \bibinfo{pages}{160--172}.
\newblock
\showISBNx{978-3-642-37456-2}


\bibitem[\protect\citeauthoryear{Cannon, Floyd, Kenyon, Walter, and
  Parry}{Cannon et~al\mbox{.}}{1997}]%
        {hyperbolic-geometry}
\bibfield{author}{\bibinfo{person}{James~W. Cannon},
  \bibinfo{person}{William~J. Floyd}, \bibinfo{person}{Richard Kenyon},
  \bibinfo{person}{Walter}, {and} \bibinfo{person}{R. Parry}.}
  \bibinfo{year}{1997}\natexlab{}.
\newblock \showarticletitle{Hyperbolic geometry}. In
  \bibinfo{booktitle}{\emph{In Flavors of geometry}}.
  \bibinfo{publisher}{University Press}, \bibinfo{pages}{59--115}.
\newblock


\bibitem[\protect\citeauthoryear{Chami, Ying, R{\'{e}}, and Leskovec}{Chami
  et~al\mbox{.}}{2019}]%
        {hgcn}
\bibfield{author}{\bibinfo{person}{Ines Chami}, \bibinfo{person}{Zhitao Ying},
  \bibinfo{person}{Christopher R{\'{e}}}, {and} \bibinfo{person}{Jure
  Leskovec}.} \bibinfo{year}{2019}\natexlab{}.
\newblock \showarticletitle{Hyperbolic Graph Convolutional Neural Networks}. In
  \bibinfo{booktitle}{\emph{Advances in Neural Information Processing Systems
  32: Annual Conference on Neural Information Processing Systems 2019, NeurIPS
  2019, December 8-14, 2019, Vancouver, BC, Canada}},
  \bibfield{editor}{\bibinfo{person}{Hanna~M. Wallach}, \bibinfo{person}{Hugo
  Larochelle}, \bibinfo{person}{Alina Beygelzimer}, \bibinfo{person}{Florence
  d'Alch{\'{e}}{-}Buc}, \bibinfo{person}{Emily~B. Fox}, {and}
  \bibinfo{person}{Roman Garnett}} (Eds.). \bibinfo{pages}{4869--4880}.
\newblock
\urldef\tempurl%
\url{https://proceedings.neurips.cc/paper/2019/hash/0415740eaa4d9decbc8da001d3fd805f-Abstract.html}
\showURL{%
\tempurl}


\bibitem[\protect\citeauthoryear{Chen, Yin, Wang, Wang, Nguyen, and Li}{Chen
  et~al\mbox{.}}{2018}]%
        {pme}
\bibfield{author}{\bibinfo{person}{Hongxu Chen}, \bibinfo{person}{Hongzhi Yin},
  \bibinfo{person}{Weiqing Wang}, \bibinfo{person}{Hao Wang},
  \bibinfo{person}{Quoc Viet~Hung Nguyen}, {and} \bibinfo{person}{Xue Li}.}
  \bibinfo{year}{2018}\natexlab{}.
\newblock \showarticletitle{{PME:} Projected Metric Embedding on Heterogeneous
  Networks for Link Prediction}. In \bibinfo{booktitle}{\emph{Proceedings of
  the 24th {ACM} {SIGKDD} International Conference on Knowledge Discovery {\&}
  Data Mining, {KDD} 2018, London, UK, August 19-23, 2018}},
  \bibfield{editor}{\bibinfo{person}{Yike Guo} {and} \bibinfo{person}{Faisal
  Farooq}} (Eds.). \bibinfo{publisher}{{ACM}}, \bibinfo{pages}{1177--1186}.
\newblock
\urldef\tempurl%
\url{https://doi.org/10.1145/3219819.3219986}
\showDOI{\tempurl}


\bibitem[\protect\citeauthoryear{Cover and Hart}{Cover and Hart}{1967}]%
        {knn}
\bibfield{author}{\bibinfo{person}{Thomas~M. Cover} {and}
  \bibinfo{person}{Peter~E. Hart}.} \bibinfo{year}{1967}\natexlab{}.
\newblock \showarticletitle{Nearest neighbor pattern classification}.
\newblock \bibinfo{journal}{\emph{{IEEE} Trans. Inf. Theory}}
  \bibinfo{volume}{13}, \bibinfo{number}{1} (\bibinfo{year}{1967}),
  \bibinfo{pages}{21--27}.
\newblock
\urldef\tempurl%
\url{https://doi.org/10.1109/TIT.1967.1053964}
\showDOI{\tempurl}


\bibitem[\protect\citeauthoryear{Duvenaud, Maclaurin, Aguilera{-}Iparraguirre,
  G{\'{o}}mez{-}Bombarelli, Hirzel, Aspuru{-}Guzik, and Adams}{Duvenaud
  et~al\mbox{.}}{2015}]%
        {molecules}
\bibfield{author}{\bibinfo{person}{David Duvenaud}, \bibinfo{person}{Dougal
  Maclaurin}, \bibinfo{person}{Jorge Aguilera{-}Iparraguirre},
  \bibinfo{person}{Rafael G{\'{o}}mez{-}Bombarelli}, \bibinfo{person}{Timothy
  Hirzel}, \bibinfo{person}{Al{\'{a}}n Aspuru{-}Guzik}, {and}
  \bibinfo{person}{Ryan~P. Adams}.} \bibinfo{year}{2015}\natexlab{}.
\newblock \showarticletitle{Convolutional Networks on Graphs for Learning
  Molecular Fingerprints}. In \bibinfo{booktitle}{\emph{Advances in Neural
  Information Processing Systems 28: Annual Conference on Neural Information
  Processing Systems 2015, December 7-12, 2015, Montreal, Quebec, Canada}},
  \bibfield{editor}{\bibinfo{person}{Corinna Cortes}, \bibinfo{person}{Neil~D.
  Lawrence}, \bibinfo{person}{Daniel~D. Lee}, \bibinfo{person}{Masashi
  Sugiyama}, {and} \bibinfo{person}{Roman Garnett}} (Eds.).
  \bibinfo{pages}{2224--2232}.
\newblock
\urldef\tempurl%
\url{https://proceedings.neurips.cc/paper/2015/hash/f9be311e65d81a9ad8150a60844bb94c-Abstract.html}
\showURL{%
\tempurl}


\bibitem[\protect\citeauthoryear{Guo, Liu, Li, Ma, Zhao, Han, Zheng, Gao, and
  Guo}{Guo et~al\mbox{.}}{2021}]%
        {hgcr}
\bibfield{author}{\bibinfo{person}{Naicheng Guo}, \bibinfo{person}{Xiaolei
  Liu}, \bibinfo{person}{Shaoshuai Li}, \bibinfo{person}{Qiongxu Ma},
  \bibinfo{person}{Yunan Zhao}, \bibinfo{person}{Bing Han},
  \bibinfo{person}{Lin Zheng}, \bibinfo{person}{Kai{-}Xin Gao}, {and}
  \bibinfo{person}{Xiaobo Guo}.} \bibinfo{year}{2021}\natexlab{}.
\newblock \showarticletitle{{HCGR:} Hyperbolic Contrastive Graph Representation
  Learning for Session-based Recommendation}.
\newblock \bibinfo{journal}{\emph{CoRR}}  \bibinfo{volume}{abs/2107.05366}
  (\bibinfo{year}{2021}).
\newblock
\showeprint[arXiv]{2107.05366}
\urldef\tempurl%
\url{https://arxiv.org/abs/2107.05366}
\showURL{%
\tempurl}


\bibitem[\protect\citeauthoryear{Hamilton, Ying, and Leskovec}{Hamilton
  et~al\mbox{.}}{2017}]%
        {graphsage}
\bibfield{author}{\bibinfo{person}{William~L. Hamilton},
  \bibinfo{person}{Zhitao Ying}, {and} \bibinfo{person}{Jure Leskovec}.}
  \bibinfo{year}{2017}\natexlab{}.
\newblock \showarticletitle{Inductive Representation Learning on Large Graphs}.
  In \bibinfo{booktitle}{\emph{Advances in Neural Information Processing
  Systems 30: Annual Conference on Neural Information Processing Systems 2017,
  December 4-9, 2017, Long Beach, CA, {USA}}},
  \bibfield{editor}{\bibinfo{person}{Isabelle Guyon}, \bibinfo{person}{Ulrike
  von Luxburg}, \bibinfo{person}{Samy Bengio}, \bibinfo{person}{Hanna~M.
  Wallach}, \bibinfo{person}{Rob Fergus}, \bibinfo{person}{S.~V.~N.
  Vishwanathan}, {and} \bibinfo{person}{Roman Garnett}} (Eds.).
  \bibinfo{pages}{1024--1034}.
\newblock
\urldef\tempurl%
\url{https://proceedings.neurips.cc/paper/2017/hash/5dd9db5e033da9c6fb5ba83c7a7ebea9-Abstract.html}
\showURL{%
\tempurl}


\bibitem[\protect\citeauthoryear{Hassani and Ahmadi}{Hassani and
  Ahmadi}{2020}]%
        {mvgcl}
\bibfield{author}{\bibinfo{person}{Kaveh Hassani} {and} \bibinfo{person}{Amir
  Hosein~Khas Ahmadi}.} \bibinfo{year}{2020}\natexlab{}.
\newblock \showarticletitle{Contrastive Multi-View Representation Learning on
  Graphs}. In \bibinfo{booktitle}{\emph{Proceedings of the 37th International
  Conference on Machine Learning, {ICML} 2020, 13-18 July 2020, Virtual Event}}
  \emph{(\bibinfo{series}{Proceedings of Machine Learning Research},
  Vol.~\bibinfo{volume}{119})}. \bibinfo{publisher}{{PMLR}},
  \bibinfo{pages}{4116--4126}.
\newblock
\urldef\tempurl%
\url{http://proceedings.mlr.press/v119/hassani20a.html}
\showURL{%
\tempurl}


\bibitem[\protect\citeauthoryear{Hjelm, Fedorov, Lavoie{-}Marchildon, Grewal,
  Bachman, Trischler, and Bengio}{Hjelm et~al\mbox{.}}{2019}]%
        {dim}
\bibfield{author}{\bibinfo{person}{R.~Devon Hjelm}, \bibinfo{person}{Alex
  Fedorov}, \bibinfo{person}{Samuel Lavoie{-}Marchildon},
  \bibinfo{person}{Karan Grewal}, \bibinfo{person}{Philip Bachman},
  \bibinfo{person}{Adam Trischler}, {and} \bibinfo{person}{Yoshua Bengio}.}
  \bibinfo{year}{2019}\natexlab{}.
\newblock \showarticletitle{Learning deep representations by mutual information
  estimation and maximization}. In \bibinfo{booktitle}{\emph{7th International
  Conference on Learning Representations, {ICLR} 2019, New Orleans, LA, USA,
  May 6-9, 2019}}. \bibinfo{publisher}{OpenReview.net}.
\newblock
\urldef\tempurl%
\url{https://openreview.net/forum?id=Bklr3j0cKX}
\showURL{%
\tempurl}


\bibitem[\protect\citeauthoryear{Kipf and Welling}{Kipf and Welling}{2017}]%
        {gcn}
\bibfield{author}{\bibinfo{person}{Thomas~N. Kipf} {and} \bibinfo{person}{Max
  Welling}.} \bibinfo{year}{2017}\natexlab{}.
\newblock \showarticletitle{Semi-Supervised Classification with Graph
  Convolutional Networks}. In \bibinfo{booktitle}{\emph{5th International
  Conference on Learning Representations, {ICLR} 2017, Toulon, France, April
  24-26, 2017, Conference Track Proceedings}}.
  \bibinfo{publisher}{OpenReview.net}.
\newblock
\urldef\tempurl%
\url{https://openreview.net/forum?id=SJU4ayYgl}
\showURL{%
\tempurl}


\bibitem[\protect\citeauthoryear{Kriege and Mutzel}{Kriege and Mutzel}{2012}]%
        {mutag}
\bibfield{author}{\bibinfo{person}{Nils~M. Kriege} {and} \bibinfo{person}{Petra
  Mutzel}.} \bibinfo{year}{2012}\natexlab{}.
\newblock \showarticletitle{Subgraph Matching Kernels for Attributed Graphs}.
  In \bibinfo{booktitle}{\emph{Proceedings of the 29th International Conference
  on Machine Learning, {ICML} 2012, Edinburgh, Scotland, UK, June 26 - July 1,
  2012}}. \bibinfo{publisher}{icml.cc / Omnipress}.
\newblock
\urldef\tempurl%
\url{http://icml.cc/2012/papers/542.pdf}
\showURL{%
\tempurl}


\bibitem[\protect\citeauthoryear{Li, Yang, Chen, and Xu}{Li
  et~al\mbox{.}}{2021b}]%
        {hncr}
\bibfield{author}{\bibinfo{person}{Anchen Li}, \bibinfo{person}{Bo Yang},
  \bibinfo{person}{Hongxu Chen}, {and} \bibinfo{person}{Guandong Xu}.}
  \bibinfo{year}{2021}\natexlab{b}.
\newblock \showarticletitle{Hyperbolic Neural Collaborative Recommender}.
\newblock \bibinfo{journal}{\emph{CoRR}}  \bibinfo{volume}{abs/2104.07414}
  (\bibinfo{year}{2021}).
\newblock
\showeprint[arXiv]{2104.07414}
\urldef\tempurl%
\url{https://arxiv.org/abs/2104.07414}
\showURL{%
\tempurl}


\bibitem[\protect\citeauthoryear{Li, Chen, Sun, Sun, Li, Cui, Yu, and Xu}{Li
  et~al\mbox{.}}{2021a}]%
        {hyperbolic-seq-rec}
\bibfield{author}{\bibinfo{person}{Yicong Li}, \bibinfo{person}{Hongxu Chen},
  \bibinfo{person}{Xiangguo Sun}, \bibinfo{person}{Zhenchao Sun},
  \bibinfo{person}{Lin Li}, \bibinfo{person}{Lizhen Cui},
  \bibinfo{person}{Philip~S. Yu}, {and} \bibinfo{person}{Guandong Xu}.}
  \bibinfo{year}{2021}\natexlab{a}.
\newblock \showarticletitle{Hyperbolic Hypergraphs for Sequential
  Recommendation}.
\newblock \bibinfo{journal}{\emph{CoRR}}  \bibinfo{volume}{abs/2108.08134}
  (\bibinfo{year}{2021}).
\newblock
\showeprint[arXiv]{2108.08134}
\urldef\tempurl%
\url{https://arxiv.org/abs/2108.08134}
\showURL{%
\tempurl}


\bibitem[\protect\citeauthoryear{Luo, Huang, Chen, Yang, and
  Baktashmotlagh}{Luo et~al\mbox{.}}{2020}]%
        {sihg}
\bibfield{author}{\bibinfo{person}{Yadan Luo}, \bibinfo{person}{Zi Huang},
  \bibinfo{person}{Hongxu Chen}, \bibinfo{person}{Yang Yang}, {and}
  \bibinfo{person}{Mahsa Baktashmotlagh}.} \bibinfo{year}{2020}\natexlab{}.
\newblock \showarticletitle{Interpretable Signed Link Prediction with Signed
  Infomax Hyperbolic Graph}.
\newblock \bibinfo{journal}{\emph{CoRR}}  \bibinfo{volume}{abs/2011.12517}
  (\bibinfo{year}{2020}).
\newblock
\showeprint[arXiv]{2011.12517}
\urldef\tempurl%
\url{https://arxiv.org/abs/2011.12517}
\showURL{%
\tempurl}


\bibitem[\protect\citeauthoryear{Nickel and Kiela}{Nickel and Kiela}{2017a}]%
        {hierarchical-representations}
\bibfield{author}{\bibinfo{person}{Maximilian Nickel} {and}
  \bibinfo{person}{Douwe Kiela}.} \bibinfo{year}{2017}\natexlab{a}.
\newblock \showarticletitle{Poincar{\'{e}} Embeddings for Learning Hierarchical
  Representations}. In \bibinfo{booktitle}{\emph{Advances in Neural Information
  Processing Systems 30: Annual Conference on Neural Information Processing
  Systems 2017, December 4-9, 2017, Long Beach, CA, {USA}}},
  \bibfield{editor}{\bibinfo{person}{Isabelle Guyon}, \bibinfo{person}{Ulrike
  von Luxburg}, \bibinfo{person}{Samy Bengio}, \bibinfo{person}{Hanna~M.
  Wallach}, \bibinfo{person}{Rob Fergus}, \bibinfo{person}{S.~V.~N.
  Vishwanathan}, {and} \bibinfo{person}{Roman Garnett}} (Eds.).
  \bibinfo{pages}{6338--6347}.
\newblock
\urldef\tempurl%
\url{https://proceedings.neurips.cc/paper/2017/hash/59dfa2df42d9e3d41f5b02bfc32229dd-Abstract.html}
\showURL{%
\tempurl}


\bibitem[\protect\citeauthoryear{Nickel and Kiela}{Nickel and Kiela}{2017b}]%
        {poincare}
\bibfield{author}{\bibinfo{person}{Maximilian Nickel} {and}
  \bibinfo{person}{Douwe Kiela}.} \bibinfo{year}{2017}\natexlab{b}.
\newblock \showarticletitle{Poincar{\'{e}} Embeddings for Learning Hierarchical
  Representations}. In \bibinfo{booktitle}{\emph{Advances in Neural Information
  Processing Systems 30: Annual Conference on Neural Information Processing
  Systems 2017, December 4-9, 2017, Long Beach, CA, {USA}}},
  \bibfield{editor}{\bibinfo{person}{Isabelle Guyon}, \bibinfo{person}{Ulrike
  von Luxburg}, \bibinfo{person}{Samy Bengio}, \bibinfo{person}{Hanna~M.
  Wallach}, \bibinfo{person}{Rob Fergus}, \bibinfo{person}{S.~V.~N.
  Vishwanathan}, {and} \bibinfo{person}{Roman Garnett}} (Eds.).
  \bibinfo{pages}{6338--6347}.
\newblock
\urldef\tempurl%
\url{https://proceedings.neurips.cc/paper/2017/hash/59dfa2df42d9e3d41f5b02bfc32229dd-Abstract.html}
\showURL{%
\tempurl}


\bibitem[\protect\citeauthoryear{Peng, Dong, Luo, Wu, and Zheng}{Peng
  et~al\mbox{.}}{2020a}]%
        {s2grl}
\bibfield{author}{\bibinfo{person}{Zhen Peng}, \bibinfo{person}{Yixiang Dong},
  \bibinfo{person}{Minnan Luo}, \bibinfo{person}{Xiao{-}Ming Wu}, {and}
  \bibinfo{person}{Qinghua Zheng}.} \bibinfo{year}{2020}\natexlab{a}.
\newblock \showarticletitle{Self-Supervised Graph Representation Learning via
  Global Context Prediction}.
\newblock \bibinfo{journal}{\emph{CoRR}}  \bibinfo{volume}{abs/2003.01604}
  (\bibinfo{year}{2020}).
\newblock
\showeprint[arXiv]{2003.01604}
\urldef\tempurl%
\url{https://arxiv.org/abs/2003.01604}
\showURL{%
\tempurl}


\bibitem[\protect\citeauthoryear{Peng, Huang, Luo, Zheng, Rong, Xu, and
  Huang}{Peng et~al\mbox{.}}{2020b}]%
        {gmi}
\bibfield{author}{\bibinfo{person}{Zhen Peng}, \bibinfo{person}{Wenbing Huang},
  \bibinfo{person}{Minnan Luo}, \bibinfo{person}{Qinghua Zheng},
  \bibinfo{person}{Yu Rong}, \bibinfo{person}{Tingyang Xu}, {and}
  \bibinfo{person}{Junzhou Huang}.} \bibinfo{year}{2020}\natexlab{b}.
\newblock \showarticletitle{Graph Representation Learning via Graphical Mutual
  Information Maximization}. In \bibinfo{booktitle}{\emph{{WWW} '20: The Web
  Conference 2020, Taipei, Taiwan, April 20-24, 2020}},
  \bibfield{editor}{\bibinfo{person}{Yennun Huang}, \bibinfo{person}{Irwin
  King}, \bibinfo{person}{Tie{-}Yan Liu}, {and} \bibinfo{person}{Maarten van
  Steen}} (Eds.). \bibinfo{publisher}{{ACM} / {IW3C2}},
  \bibinfo{pages}{259--270}.
\newblock
\urldef\tempurl%
\url{https://doi.org/10.1145/3366423.3380112}
\showDOI{\tempurl}


\bibitem[\protect\citeauthoryear{Qiu, Chen, Dong, Zhang, Yang, Ding, Wang, and
  Tang}{Qiu et~al\mbox{.}}{2020}]%
        {gcc}
\bibfield{author}{\bibinfo{person}{Jiezhong Qiu}, \bibinfo{person}{Qibin Chen},
  \bibinfo{person}{Yuxiao Dong}, \bibinfo{person}{Jing Zhang},
  \bibinfo{person}{Hongxia Yang}, \bibinfo{person}{Ming Ding},
  \bibinfo{person}{Kuansan Wang}, {and} \bibinfo{person}{Jie Tang}.}
  \bibinfo{year}{2020}\natexlab{}.
\newblock \showarticletitle{{GCC:} Graph Contrastive Coding for Graph Neural
  Network Pre-Training}. In \bibinfo{booktitle}{\emph{{KDD} '20: The 26th {ACM}
  {SIGKDD} Conference on Knowledge Discovery and Data Mining, Virtual Event,
  CA, USA, August 23-27, 2020}}, \bibfield{editor}{\bibinfo{person}{Rajesh
  Gupta}, \bibinfo{person}{Yan Liu}, \bibinfo{person}{Jiliang Tang}, {and}
  \bibinfo{person}{B.~Aditya Prakash}} (Eds.). \bibinfo{publisher}{{ACM}},
  \bibinfo{pages}{1150--1160}.
\newblock
\urldef\tempurl%
\url{https://doi.org/10.1145/3394486.3403168}
\showDOI{\tempurl}


\bibitem[\protect\citeauthoryear{Rozemberczki and Sarkar}{Rozemberczki and
  Sarkar}{2020}]%
        {diffusion-sampler}
\bibfield{author}{\bibinfo{person}{Benedek Rozemberczki} {and}
  \bibinfo{person}{Rik Sarkar}.} \bibinfo{year}{2020}\natexlab{}.
\newblock \showarticletitle{Fast Sequence-Based Embedding with Diffusion
  Graphs}.
\newblock \bibinfo{journal}{\emph{CoRR}}  \bibinfo{volume}{abs/2001.07463}
  (\bibinfo{year}{2020}).
\newblock
\showeprint[arxiv]{2001.07463}
\urldef\tempurl%
\url{https://arxiv.org/abs/2001.07463}
\showURL{%
\tempurl}


\bibitem[\protect\citeauthoryear{Song, Chen, Liu, Jiang, and Wang}{Song
  et~al\mbox{.}}{2021}]%
        {hyper-node-emb}
\bibfield{author}{\bibinfo{person}{Wenzhuo Song}, \bibinfo{person}{Hongxu
  Chen}, \bibinfo{person}{Xueyan Liu}, \bibinfo{person}{Hongzhe Jiang}, {and}
  \bibinfo{person}{Shengsheng Wang}.} \bibinfo{year}{2021}\natexlab{}.
\newblock \showarticletitle{Hyperbolic node embedding for signed networks}.
\newblock \bibinfo{journal}{\emph{Neurocomputing}}  \bibinfo{volume}{421}
  (\bibinfo{year}{2021}), \bibinfo{pages}{329--339}.
\newblock
\urldef\tempurl%
\url{https://doi.org/10.1016/j.neucom.2020.10.008}
\showDOI{\tempurl}


\bibitem[\protect\citeauthoryear{van~den Berg, Kipf, and Welling}{van~den Berg
  et~al\mbox{.}}{2017}]%
        {matrix-completion}
\bibfield{author}{\bibinfo{person}{Rianne van~den Berg},
  \bibinfo{person}{Thomas~N. Kipf}, {and} \bibinfo{person}{Max Welling}.}
  \bibinfo{year}{2017}\natexlab{}.
\newblock \showarticletitle{Graph Convolutional Matrix Completion}.
\newblock \bibinfo{journal}{\emph{CoRR}}  \bibinfo{volume}{abs/1706.02263}
  (\bibinfo{year}{2017}).
\newblock
\showeprint[arXiv]{1706.02263}
\urldef\tempurl%
\url{http://arxiv.org/abs/1706.02263}
\showURL{%
\tempurl}


\bibitem[\protect\citeauthoryear{Velickovic, Cucurull, Casanova, Romero,
  Li{\`{o}}, and Bengio}{Velickovic et~al\mbox{.}}{2018}]%
        {gat}
\bibfield{author}{\bibinfo{person}{Petar Velickovic}, \bibinfo{person}{Guillem
  Cucurull}, \bibinfo{person}{Arantxa Casanova}, \bibinfo{person}{Adriana
  Romero}, \bibinfo{person}{Pietro Li{\`{o}}}, {and} \bibinfo{person}{Yoshua
  Bengio}.} \bibinfo{year}{2018}\natexlab{}.
\newblock \showarticletitle{Graph Attention Networks}. In
  \bibinfo{booktitle}{\emph{6th International Conference on Learning
  Representations, {ICLR} 2018, Vancouver, BC, Canada, April 30 - May 3, 2018,
  Conference Track Proceedings}}. \bibinfo{publisher}{OpenReview.net}.
\newblock
\urldef\tempurl%
\url{https://openreview.net/forum?id=rJXMpikCZ}
\showURL{%
\tempurl}


\bibitem[\protect\citeauthoryear{Velickovic, Fedus, Hamilton, Li{\`{o}},
  Bengio, and Hjelm}{Velickovic et~al\mbox{.}}{2019}]%
        {dgi}
\bibfield{author}{\bibinfo{person}{Petar Velickovic}, \bibinfo{person}{William
  Fedus}, \bibinfo{person}{William~L. Hamilton}, \bibinfo{person}{Pietro
  Li{\`{o}}}, \bibinfo{person}{Yoshua Bengio}, {and} \bibinfo{person}{R.~Devon
  Hjelm}.} \bibinfo{year}{2019}\natexlab{}.
\newblock \showarticletitle{Deep Graph Infomax}. In
  \bibinfo{booktitle}{\emph{7th International Conference on Learning
  Representations, {ICLR} 2019, New Orleans, LA, USA, May 6-9, 2019}}.
  \bibinfo{publisher}{OpenReview.net}.
\newblock
\urldef\tempurl%
\url{https://openreview.net/forum?id=rklz9iAcKQ}
\showURL{%
\tempurl}


\bibitem[\protect\citeauthoryear{Wang, Min, Chen, and Wu}{Wang
  et~al\mbox{.}}{2021}]%
        {drug-discovery}
\bibfield{author}{\bibinfo{person}{Yingheng Wang}, \bibinfo{person}{Yaosen
  Min}, \bibinfo{person}{Xin Chen}, {and} \bibinfo{person}{Ji Wu}.}
  \bibinfo{year}{2021}\natexlab{}.
\newblock \showarticletitle{Multi-view Graph Contrastive Representation
  Learning for Drug-Drug Interaction Prediction}. In
  \bibinfo{booktitle}{\emph{{WWW} '21: The Web Conference 2021, Virtual Event /
  Ljubljana, Slovenia, April 19-23, 2021}},
  \bibfield{editor}{\bibinfo{person}{Jure Leskovec}, \bibinfo{person}{Marko
  Grobelnik}, \bibinfo{person}{Marc Najork}, \bibinfo{person}{Jie Tang}, {and}
  \bibinfo{person}{Leila Zia}} (Eds.). \bibinfo{publisher}{{ACM} / {IW3C2}},
  \bibinfo{pages}{2921--2933}.
\newblock
\urldef\tempurl%
\url{https://doi.org/10.1145/3442381.3449786}
\showDOI{\tempurl}


\bibitem[\protect\citeauthoryear{Xu, Hu, Leskovec, and Jegelka}{Xu
  et~al\mbox{.}}{2019}]%
        {gin}
\bibfield{author}{\bibinfo{person}{Keyulu Xu}, \bibinfo{person}{Weihua Hu},
  \bibinfo{person}{Jure Leskovec}, {and} \bibinfo{person}{Stefanie Jegelka}.}
  \bibinfo{year}{2019}\natexlab{}.
\newblock \showarticletitle{How Powerful are Graph Neural Networks?}. In
  \bibinfo{booktitle}{\emph{7th International Conference on Learning
  Representations, {ICLR} 2019, New Orleans, LA, USA, May 6-9, 2019}}.
  \bibinfo{publisher}{OpenReview.net}.
\newblock
\urldef\tempurl%
\url{https://openreview.net/forum?id=ryGs6iA5Km}
\showURL{%
\tempurl}


\bibitem[\protect\citeauthoryear{Yanardag and Vishwanathan}{Yanardag and
  Vishwanathan}{2015}]%
        {graph-kernels}
\bibfield{author}{\bibinfo{person}{Pinar Yanardag} {and}
  \bibinfo{person}{S.~V.~N. Vishwanathan}.} \bibinfo{year}{2015}\natexlab{}.
\newblock \showarticletitle{Deep Graph Kernels}. In
  \bibinfo{booktitle}{\emph{Proceedings of the 21th {ACM} {SIGKDD}
  International Conference on Knowledge Discovery and Data Mining, Sydney, NSW,
  Australia, August 10-13, 2015}}, \bibfield{editor}{\bibinfo{person}{Longbing
  Cao}, \bibinfo{person}{Chengqi Zhang}, \bibinfo{person}{Thorsten Joachims},
  \bibinfo{person}{Geoffrey~I. Webb}, \bibinfo{person}{Dragos~D. Margineantu},
  {and} \bibinfo{person}{Graham Williams}} (Eds.). \bibinfo{publisher}{{ACM}},
  \bibinfo{pages}{1365--1374}.
\newblock
\urldef\tempurl%
\url{https://doi.org/10.1145/2783258.2783417}
\showDOI{\tempurl}


\bibitem[\protect\citeauthoryear{Yang, Chen, Li, Yu, and Xu}{Yang
  et~al\mbox{.}}{2021}]%
        {hmg-cr}
\bibfield{author}{\bibinfo{person}{Haoran Yang}, \bibinfo{person}{Hongxu Chen},
  \bibinfo{person}{Lin Li}, \bibinfo{person}{Philip~S. Yu}, {and}
  \bibinfo{person}{Guandong Xu}.} \bibinfo{year}{2021}\natexlab{}.
\newblock \showarticletitle{Hyper Meta-Path Contrastive Learning for
  Multi-Behavior Recommendation}.
\newblock \bibinfo{journal}{\emph{CoRR}}  \bibinfo{volume}{abs/2109.02859}
  (\bibinfo{year}{2021}).
\newblock
\showeprint[arXiv]{2109.02859}
\urldef\tempurl%
\url{https://arxiv.org/abs/2109.02859}
\showURL{%
\tempurl}


\bibitem[\protect\citeauthoryear{You, Chen, Sui, Chen, Wang, and Shen}{You
  et~al\mbox{.}}{2020}]%
        {graph-aug}
\bibfield{author}{\bibinfo{person}{Yuning You}, \bibinfo{person}{Tianlong
  Chen}, \bibinfo{person}{Yongduo Sui}, \bibinfo{person}{Ting Chen},
  \bibinfo{person}{Zhangyang Wang}, {and} \bibinfo{person}{Yang Shen}.}
  \bibinfo{year}{2020}\natexlab{}.
\newblock \showarticletitle{Graph Contrastive Learning with Augmentations}. In
  \bibinfo{booktitle}{\emph{Advances in Neural Information Processing Systems
  33: Annual Conference on Neural Information Processing Systems 2020, NeurIPS
  2020, December 6-12, 2020, virtual}}, \bibfield{editor}{\bibinfo{person}{Hugo
  Larochelle}, \bibinfo{person}{Marc'Aurelio Ranzato}, \bibinfo{person}{Raia
  Hadsell}, \bibinfo{person}{Maria{-}Florina Balcan}, {and}
  \bibinfo{person}{Hsuan{-}Tien Lin}} (Eds.).
\newblock
\urldef\tempurl%
\url{https://proceedings.neurips.cc/paper/2020/hash/3fe230348e9a12c13120749e3f9fa4cd-Abstract.html}
\showURL{%
\tempurl}


\bibitem[\protect\citeauthoryear{Zhang, Chen, Ming, Cui, Yin, and Xu}{Zhang
  et~al\mbox{.}}{2021}]%
        {embedding-space}
\bibfield{author}{\bibinfo{person}{Sixiao Zhang}, \bibinfo{person}{Hongxu
  Chen}, \bibinfo{person}{Xiao Ming}, \bibinfo{person}{Lizhen Cui},
  \bibinfo{person}{Hongzhi Yin}, {and} \bibinfo{person}{Guandong Xu}.}
  \bibinfo{year}{2021}\natexlab{}.
\newblock \showarticletitle{Where are we in embedding spaces? {A} Comprehensive
  Analysis on Network Embedding Approaches for Recommender Systems}.
\newblock \bibinfo{journal}{\emph{CoRR}}  \bibinfo{volume}{abs/2105.08908}
  (\bibinfo{year}{2021}).
\newblock
\showeprint[arxiv]{2105.08908}
\urldef\tempurl%
\url{https://arxiv.org/abs/2105.08908}
\showURL{%
\tempurl}


\bibitem[\protect\citeauthoryear{Zhu, Xu, Yu, Liu, Wu, and Wang}{Zhu
  et~al\mbox{.}}{2021}]%
        {gca}
\bibfield{author}{\bibinfo{person}{Yanqiao Zhu}, \bibinfo{person}{Yichen Xu},
  \bibinfo{person}{Feng Yu}, \bibinfo{person}{Qiang Liu}, \bibinfo{person}{Shu
  Wu}, {and} \bibinfo{person}{Liang Wang}.} \bibinfo{year}{2021}\natexlab{}.
\newblock \showarticletitle{Graph Contrastive Learning with Adaptive
  Augmentation}. In \bibinfo{booktitle}{\emph{{WWW} '21: The Web Conference
  2021, Virtual Event / Ljubljana, Slovenia, April 19-23, 2021}},
  \bibfield{editor}{\bibinfo{person}{Jure Leskovec}, \bibinfo{person}{Marko
  Grobelnik}, \bibinfo{person}{Marc Najork}, \bibinfo{person}{Jie Tang}, {and}
  \bibinfo{person}{Leila Zia}} (Eds.). \bibinfo{publisher}{{ACM} / {IW3C2}},
  \bibinfo{pages}{2069--2080}.
\newblock
\urldef\tempurl%
\url{https://doi.org/10.1145/3442381.3449802}
\showDOI{\tempurl}


\end{thebibliography}

\clearpage
\appendix

\section{Hyper-parameter Settings}
\begin{table*}[hb]
\centering
\caption{Hyper-parameter settings of the proposed DSGC in the comparison study on all the datasets with different ratio of labeled data in training set.}
\label{tab: hyper-parameter}
\resizebox{\textwidth}{24mm}{
\begin{tabular}{c|ccc|ccc|ccc}
\toprule
\multirow{2}{*}{\diagbox{Hyper-parameters}{Dataset \& Label Ratio}} & \multicolumn{3}{c|}{MUATG} & \multicolumn{3}{c|}{REDDIT-BINARY} & \multicolumn{3}{c}{COLLAB}        \\ \cmidrule{2-10} 
                                                                          & 0.1        & 0.3    & 0.5  & 0.1        & 0.3       & 0.5       & 0.1       & 0.3       & 0.5       \\ \midrule
Euclidean Encoder                                                         & GraphSAGE  & GCN    & GCN  & GCN        & GCN       & GIN       & GraphSAGE & GraphSAGE & GraphSAGE \\
Hyperbolic Encoder                                                        & GraphSAGE  & GCN    & GCIN & GAT        & GIN       & GIN       & GCN       & GraphSAGE & GIN       \\
Number of Encoder Layers                                                  & 3          & 3      & 3    & 1          & 1         & 1         & 3         & 3         & 1         \\
Temperature                                                               & 1          & 1      & 1    & 100        & 100       & 100       & 100       & 100       & 100       \\
Learning Rate                                                             & 1e-4       & 1.7e-4 & 5e-5 & 2e-5       & 1e-4      & 1e-5      & 2e-5      & 2e-5      & 2e-5      \\
Weight Decay                                                              & 1e-5       & 1e-5   & 1e-5 & 1e-5       & 1e-5      & 1e-5      & 1e-5      & 1e-5      & 1e-5      \\
Number of Training Epoch                                                  & 200        & 200    & 200  & 200        & 200       & 200       & 200       & 200       & 200       \\
Weight of Contrastive Learning   (w)                                      & 0.01       & 0.01   & 0.01 & 1e-5       & 0.01      & 0.01      & 0.01      & 1e-4      & 0.01      \\
Batch Size                                                                & 8          & 8      & 8    & 16         & 16        & 16        & 32        & 64        & 64        \\
Hidden Dimension                                                          & 16         & 16     & 16   & 16         & 16        & 16        & 16        & 16        & 16        \\ \bottomrule
\end{tabular}
}
\end{table*}

For the reproducibility of our work, we list all the hyper-parameter settings of comparison study in Table \ref{tab: hyper-parameter}.

\end{document}